\documentclass[11pt]{article}

\usepackage[margin=1in]{geometry}
\usepackage{framed}
\usepackage{multirow}
\usepackage{amsmath}
\usepackage{enumitem}
\usepackage{fancybox}
\usepackage{mathtools}
\usepackage{amsthm}
\usepackage{color}
\usepackage[most]{tcolorbox}
\usepackage{caption}
\usepackage{subcaption}
\usepackage{amsfonts}
\usepackage{algorithm}
\usepackage{algpseudocode}
\usepackage{xcolor}
\usepackage{natbib}
\usepackage{threeparttable}
\usepackage[table]{xcolor}
\usepackage{booktabs}
\usepackage{graphicx}
\usepackage{hyperref}
\usepackage{url}

\DeclareMathAlphabet{\bbold}{U}{bbold}{m}{n}

\newtheorem{problem}{Problem}

\newtheorem{remark}{Remark}
\newtheorem{definition}{Definition}
\newtheorem{proposition}{Proposition}

\newtcolorbox[auto counter, number within=section]{examplebox}
{colframe=black,
 colback=white,
 coltitle=black,
 arc=1mm,
 boxrule=0.2mm,
 toptitle=1mm,
 bottomtitle=1mm,
 boxsep=0.6mm,
  left=2.2mm,
  right=2.2mm,
  top=0.5mm,
  bottom=0.5mm
 }

\title{Beyond Independent Manipulation: Individual Fairness-aware Strategic Classification with Peer Imitation}

\author{
Xinpeng Lv$^1$ \quad
Yunxin Mao$^1$ \quad
Renzhe Xu$^2$ \quad
Jinxuan Yang$^4$ \quad
Chunyuan Zheng$^3$ \\
Yuanlong Chen$^5$ \quad
Wangrong Huang$^1$ \quad
Shaowu Yang$^1$ \quad
Wenjing Yang$^1$ \quad
Xinwang Liu$^1$ \\
Peng Cui$^6$ \quad
Haotian Wang$^{1\dagger}$
\\
$^1$College of Computer Science and Technology, National University of Defense Technology \\
$^3$School of Mathematical Sciences, Peking University \\
$^2$Institute for Theoretical Computer Science, Shanghai University of Finance and Economics \\
$^4$Information Technology Development, Aetos Capital Group, Sydney \\
$^5$Faculty of Computing, Harbin Institute of Technology \\
$^6$Department of Computer Science and Technology, Tsinghua University \\
\\
{\small \texttt{\{lvxinpeng,wanghaotian13\}@nudt.edu.cn} \quad $^\dagger$Corresponding author}
}

\date{}

\begin{document}

\maketitle

\begin{abstract}
Strategic classification~(SC) investigates scenarios where agents manipulate their features to obtain favorable decisions from predictive models.
Existing fairness-aware SC approaches primarily focus on group fairness and typically assume that agents respond independently.
However, when individual fairness is required, ensuring similar individuals receive similar outcomes, agents' manipulation becomes interdependent: an agent's preferred manipulation depends on the neighborhoods' outcomes.
This induces a mismatch between classical SC formulations and fairness-aware decision settings, where independent models no longer accurately characterize strategic manipulations.
To address this issue, we introduce \textbf{individual fairness-aware strategic classification}~\textit{(IFSC)}, a framework that models peer-driven manipulation arising from individual fairness, where agents imitate nearby positively decided peers to obtain favorable outcomes. IFSC characterizes strategic manipulation as similarity-based imitation toward visible accepted peers and learns classifiers under the resulting post-manipulation distributions. To account for uncertainty in peer observability, IFSC employs a robust learning process that introduces stochastic perturbations during manipulation simulation.
Experiments on synthetic and real-world datasets demonstrate that IFSC improves individual-fairness consistency and mitigates imitation-induced distortions.
\end{abstract}

\section{Introduction}

Machine learning models are increasingly deployed to support decision-making in domains such as hiring~\cite{sanchez2020does}, credit scoring~\cite{jagtiani2019roles}, and college admissions~\cite{kuvcak2018machine}.
In such settings, individuals are not passive recipients of decision outcomes.
Instead, they may strategically adjust their \emph{observable} features to obtain more favorable decisions, thereby challenging the reliability of the deployed classifier.
This phenomenon is often summarized by Goodhart's Law~\cite{Strathern_1997}: \textit{when a measure becomes a target, it ceases to be a good measure}.
For example, a loan applicant may temporarily inflate reported income to appear more creditworthy.

To explicitly account for strategic manipulation, strategic classification (SC)~\cite{hardt2016strategic} models the interaction between a classifier and strategic agents, typically as a Stackelberg game in which the decision maker commits to a classification rule and individuals respond by manipulating features subject to cost~\cite{pmlr-v139-ghalme21a,singh2024optimal,chen2020learning}.
However, robustness alone is insufficient in real-world decision settings where fairness is a central requirement.
Since decisions in domains such as hiring or lending directly affect individual opportunities, strategic behavior may amplify existing inequities rather than mitigate them~\cite{fairnessreverse}. For instance, historical biases may still lead to systematically worse outcomes for certain groups even when strategic behavior is accounted for~\cite{zemel2013learning,zhang2022fairness}.
As a result, prior work has incorporated fairness considerations into SC primarily through group fairness notions, including demographic parity~\cite{zemel2013learning}, equality of opportunity~\cite{roemer2015equality}, and predictive parity~\cite{dieterich2016compas}, typically enforced via group-dependent constraints~\cite{zhang2022fairness,shimao2025strategic}.

However, group fairness alone may be insufficient to address unfairness in strategic environments. Group fairness focuses on achieving statistical parity across groups, but does not guarantee local consistency across individuals. In strategic environments, such inconsistencies introduce additional unfairness, as similar individuals may face different opportunities to obtain favorable outcomes, a phenomenon that we analyze in detail in Sec.~\ref{sec:if_needed}.
For example, two applicants with nearly identical financial profiles but belonging to different groups may receive opposite decisions, where one is accepted directly while the other must incur additional effort or strategic manipulation to achieve the same outcome.
This motivates a shift from group-level fairness to individual fairness, which requires that individuals who are similar receive similar decisions.

\begin{figure}[t]
    \centering
    \includegraphics[width=0.98\linewidth]{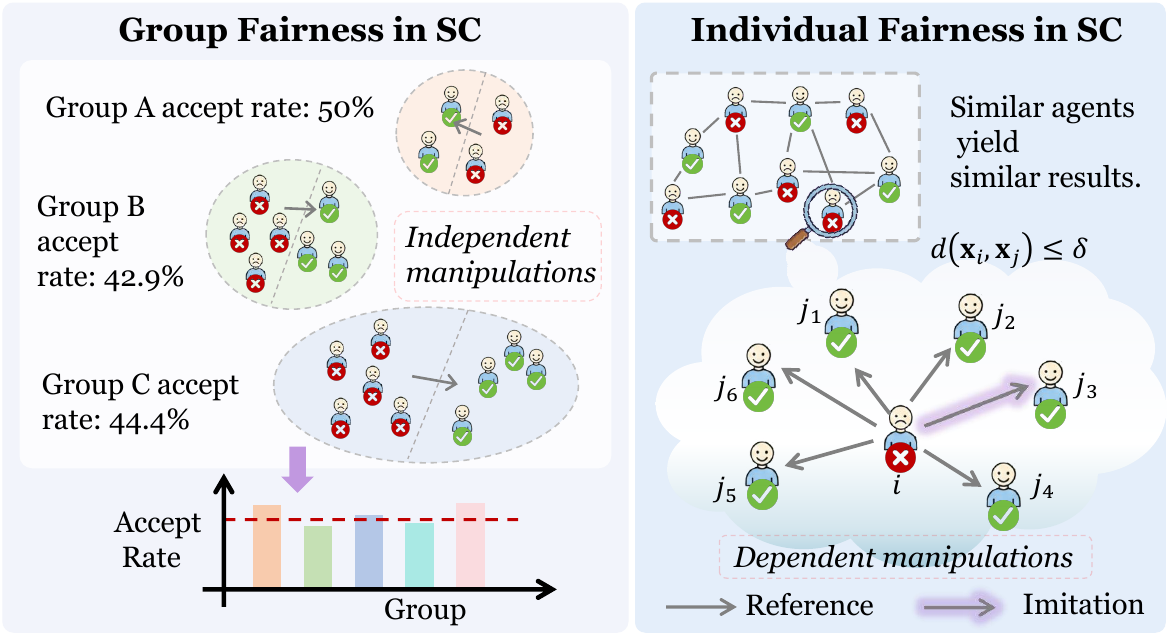}
    \vskip -0.1in
    \caption{Comparison between group fairness and individual fairness in strategic classification. Group fairness preserves independent manipulation, while individual fairness induces similarity-based interdependent strategic manipulation.}
    \label{fig1}
    \vskip -0.2in
\end{figure}

Unfortunately, introducing individual fairness into strategic classification challenges the key structural assumption underlying classical SC formulations.
Standard SC models assume that agents make strategic manipulations independently.
Under individual fairness, however, an agent's optimal manipulation often depends on the outcomes of its neighborhood, leading to interdependent manipulation, with theoretical analysis in detail in Sec.~\ref{sec:ifsc_breakdown}.
Specifically, agents may improve their outcomes by imitating positively decided peers within their similarity neighborhood, leading to locally interdependent strategic manipulation.
As illustrated in Figure~\ref{fig1}, while agents under group fairness exhibit independent manipulation, agents under individual fairness tend to imitate similar peers, making strategic manipulation locally dependent.
This motivates the following question:

\begin{examplebox}
\textit{\;\;How can strategic classification be reformulated to capture interdependent manipulations induced by individual fairness?}
\end{examplebox}
\vskip -0.05in

To address this challenge, we propose \emph{individual fairness-aware strategic classification (IFSC)}, a framework designed for strategic settings where agents imitate nearby positively decided peers to obtain similar decisions under individual fairness.
Concretely, IFSC models strategic manipulation under individual fairness as peer imitation: each agent observes a set of visible positively decided peers and strategically manipulates its features to become similar to these peers in order to obtain a favorable decision.
To account for such manipulation, IFSC introduces a robust learning process that adds stochastic perturbations to agents' visible peer sets and optimizes the classifier on the resulting post-manipulation distributions.
Our contributions are summarized as follows:

\begin{itemize}[leftmargin=18pt]
    \item We identify a fundamental limitation of classical strategic classification under individual fairness: the standard assumption of agent-independent manipulation no longer holds when agents exploit similarity neighborhoods.
    We show that individual fairness constraints can induce interdependent strategic manipulation through peer imitation, as agents strategically move toward the neighborhoods of nearby positively decided peers, a phenomenon not captured by existing SC models.

    \item We propose \textbf{individual-fairness-aware strategic classification}~\emph{(IFSC)}, a strategic framework that models peer-driven manipulation, where agents strategically move toward the neighborhoods of visible positively decided peers.
    Building on this structure, IFSC introduces a principled training objective that optimizes classifiers over simulated post-manipulation distributions with stochastic perturbations, thereby aligning learning with the induced strategic distribution.

    \item We provide extensive empirical evaluation on both synthetic and real-world datasets, demonstrating that IFSC consistently preserves individual fairness consistency and post-manipulation accuracy.
    The results show that accounting for fairness-induced strategic dependence leads to more robust decisions under fairness-aware strategic responses while maintaining competitive predictive performance.
\end{itemize}

\section{Related Work}
\label{rwork}
\subsection{Strategic Machine Learning}
Strategic classification~\cite{hardt2016strategic} studies settings where individuals manipulate their features to influence model outcomes~\cite{dong2017strategicclassificationrevealedpreferences,shavit2020causal,chen2020learning,NEURIPS2021_f1404c26,zrnic2021leads,tsirtsis2024optimal,lv2026breaking,lv2026tabular}. Several works address challenges arising from unknown manipulations or limited agent information~\cite{shao2024strategic,pmlr-v139-ghalme21a,lv2026beyond}.
Recent studies incorporate causal reasoning into strategic learning~\cite{miller2020strategic,chen2023learning,horowitz2023causal,vo2024causal,efthymiou2025incentivizing,chang2024s}, distinguishing between manipulable and genuinely improvable features. This line of work emphasizes how strategic responses may reflect or distort true underlying qualifications.
Relatedly, performative prediction~\cite{perdomo2020performative,rosenfeld2020predictions,hardt2022performative,hardt2023performative,mendler2022anticipating,mofakhami2023performative} analyzes how predictive models influence the data distribution over time through repeated deployment.
A complementary direction focuses on promoting social welfare~\cite{haghtalab2020maximizing,estornell2023incentivizing,xie2024non}, designing mechanisms that align agent incentives with collective benefit.

\subsection{Fairness in Strategic Classification}
Recent research has examined the complex interplay between fairness and strategic behavior in machine learning. Several studies highlight that unequal manipulation costs can exacerbate disparities even in the presence of fairness constraints~\cite{hu2019disparate,milli2019social}.
To mitigate these issues, various methods have been proposed, such as optimizing classifiers to reduce manipulation costs for disadvantaged groups~\cite{keswani2023addressing} or employing minimax group fairness frameworks~\cite{diana2024minimax}. Other work evaluates fairness through agents' equilibrium behaviors~\cite{shimao2025strategic} and explores how incentive structures can influence manipulation~\cite{zhang2022fairness}. Moreover, group fairness constraints may unintentionally result in fairness reversal when agents strategically modify their features~\cite{fairnessreverse,lv2026partial}.
Recent efforts also focus on constructing fairness-aware models that anticipate manipulation and promote genuine improvement~\cite{alhanouti2025anticipating}. Another work~\cite{zuoindividual} focuses on how to achieve individually fair decisions under strategic behavior. In contrast, our work examines a different effect of individual fairness: once similar individuals are expected to receive similar outcomes, agents may exploit local accepted peers as references, making manipulation population-dependent rather than independent.

\subsection{Individual Fairness}
Individual fairness formalizes the principle that \emph{similar individuals should receive similar decisions}, typically defined with respect to a task-specific similarity metric over the feature space~\cite{dwork2012fairness}.
Unlike group fairness, individual fairness imposes local consistency requirements on the decision rule at the level of individual pairs or neighborhoods.
A substantial body of work studies how to operationalize and enforce individual fairness in machine learning models, including specifying similarity metrics~\cite{ilvento2020metric,mukherjee2020two}, designing learning frameworks that approximately satisfy metric-based fairness constraints~\cite{yona2018probably}, and developing representation learning or post-processing methods to promote locally consistent decisions~\cite{lahoti2019ifair}.
More recent studies extend individual fairness to sequential decision-making settings, such as online learning and bandits~\cite{joseph2016fairness,gupta2021individual}.
In parallel, recent work revisits how individual fairness should be \emph{measured} beyond simple consistency-based criteria~\cite{waller2025beyond}, and develops \emph{formal certification} or \emph{correct-by-construction} training methods that provide provable fairness guarantees for deep models~\cite{wicker2023distributional,zhang2025correct}.

\section{Preliminaries}

We briefly review strategic classification and fairness notions used in this work.
Throughout, random variables are denoted by uppercase letters (e.g., $X,Y$) and their realizations by lowercase letters (e.g., $x,y$).
Bold symbols (e.g., $\mathbf{x},\mathbf{X}$) are used for vectors or matrices.

\subsection{Strategic Classification}

Strategic classification is typically modeled as a Stackelberg game~\cite{li2017review,hardt2016strategic,miller2020strategic} between a \emph{decision maker} and a population of \emph{decision subjects} (agents).
The decision maker deploys a classifier
\begin{equation}
    f : \mathcal{X} \subseteq \mathbb{R}^d \to \{0,1\},
\end{equation}
while each agent with features $\mathbf{x} \in \mathcal{X}$ may strategically modify them to $\mathbf{x}' \in \mathcal{X}$ at some cost $c(\mathbf{x},\mathbf{x}') \ge 0$. The agent's behavior is captured by a best-response map that maximizes a utility trade-off between the classification outcome and manipulation cost.

\begin{definition}[Strategic manipulation]
Given a deployed classifier $f$ and cost function $c:\mathcal{X}\times\mathcal{X}\to\mathbb{R}_{\ge 0}$, the agent with original features $\mathbf{x}$ chooses a manipulated feature
\begin{equation}
    \mathbf{x}' \;=\; b(\mathbf{x};f)
    \;\in\; \arg\max_{\tilde{\mathbf{x}} \in \mathcal{X}}
    U(\mathbf{x},\tilde{\mathbf{x}};f),
    \label{eq:best_response}
\end{equation}
where the utility is
\begin{equation}
    U(\mathbf{x},\tilde{\mathbf{x}};f)
    \;=\; f(\tilde{\mathbf{x}}) \;-\; \lambda\, c(\mathbf{x},\tilde{\mathbf{x}}),
    \label{eq:utility}
\end{equation}
and $\lambda>0$ is a trade-off parameter between the benefit of receiving a positive decision and the cost of manipulation.
\end{definition}

The decision maker anticipates such strategic behavior and chooses a decision rule that optimizes predictive performance under the induced post-manipulation distribution.

\begin{definition}[Decision optimization under strategic behavior]
Let $(\mathbf{x},y)\sim\mathcal{D}$ denote the distribution of original features and labels.
Given a hypothesis class $\mathcal{F}$, the decision maker solves
\begin{equation}
    f^\star \;\in\;
    \arg\max_{f \in \mathcal{F}}
    \;\mathbf{E}_{(\mathbf{x},y)\sim\mathcal{D}}
    \Big[ \bbold{1}\big( f\big(b(\mathbf{x};f)\big) = y \big) \Big],
    \label{eq:sc_outer}
\end{equation}
where $b(\mathbf{x};f)$ is the best-response function in~\eqref{eq:best_response}.
\end{definition}

\subsection{Fairness in Strategic Classification}

We now recall how fairness notions are incorporated into the strategic classification setting.

\noindent \textbf{Group fairness.} Let $(\mathbf{x},y,g)\sim\mathcal{D}$ denote the distribution of features, labels, and group memberships.
Given a deployed decision rule $f$, each agent chooses a post-manipulation feature $\mathbf{x}'$ according to the manipulation function in~\eqref{eq:best_response}.
To accommodate group-dependent policies, the induced manipulation also depends on the group attribute $g$, and we write
\begin{equation}
    \mathbf{x}' = b_g(\mathbf{x}; g, f).
\end{equation}
\noindent Therefore, group fairness is typically imposed through constraints on statistics of post-manipulation predictions.
A generic formulation can be written as
\begin{equation}
    \mathcal{C}(f;\mathcal{D},G)
    \;=\;
    \Phi\Big( \big\{ \Pr\big(f(\mathbf{x}')=1 \mid g\big) : g\in G \big\} \Big)
    \;\le\; \varepsilon,
    \label{eq:group_fair_sc}
\end{equation}
where $G$ is the set of groups, $\Phi$ is a disparity functional (e.g., the maximum difference of positive prediction rates), and $\varepsilon\ge 0$ is a tolerance.

\noindent \textbf{Individual fairness.}
Individual fairness requires that \emph{similar individuals} receive similar decisions.
Given a similarity metric $d:\mathcal{X}\times\mathcal{X}\to\mathbb{R}_{\ge 0}$ and a similarity radius $\delta>0$, an individually fair classifier is expected to satisfy, after manipulation,
\begin{equation}
    d(\mathbf{x}_i,\mathbf{x}_j) \;\le\; \delta
    \quad\Rightarrow\quad
    f(\mathbf{x}_i) \;=\; f(\mathbf{x}_j),
    \label{eq:if_sc_post}
\end{equation}
for all agents $i,j$.

\noindent {\bf Divergence between two fairness constraints.} Unlike group fairness, which constrains aggregate statistics while preserving independent manipulations, individual fairness alters the structure of manipulation.
In particular, an agent's manipulation depends on nearby individuals $\mathbf{x}_j$ to obtain a consistent decision, which can be expressed as
\begin{equation}
    \mathbf{x}'_i = b_{\mathrm{IF}}\!\left(\mathbf{x}_i; f, \mathbf{x}_j\right),
    \quad
    \text{with } d(\mathbf{x}'_i,\mathbf{x}_j)\le \delta,
    \label{eq:if_response}
\end{equation}
indicating that post-manipulation features are chosen relative to similar individuals rather than independently with respect to the decision rule alone. We will analyze this difference with its consequences in the later section.

\section{Problem Statement}

\begin{figure*}[t]
    \centering
    \begin{subfigure}[t]{0.33\textwidth}
        \centering
        \includegraphics[width=\linewidth]{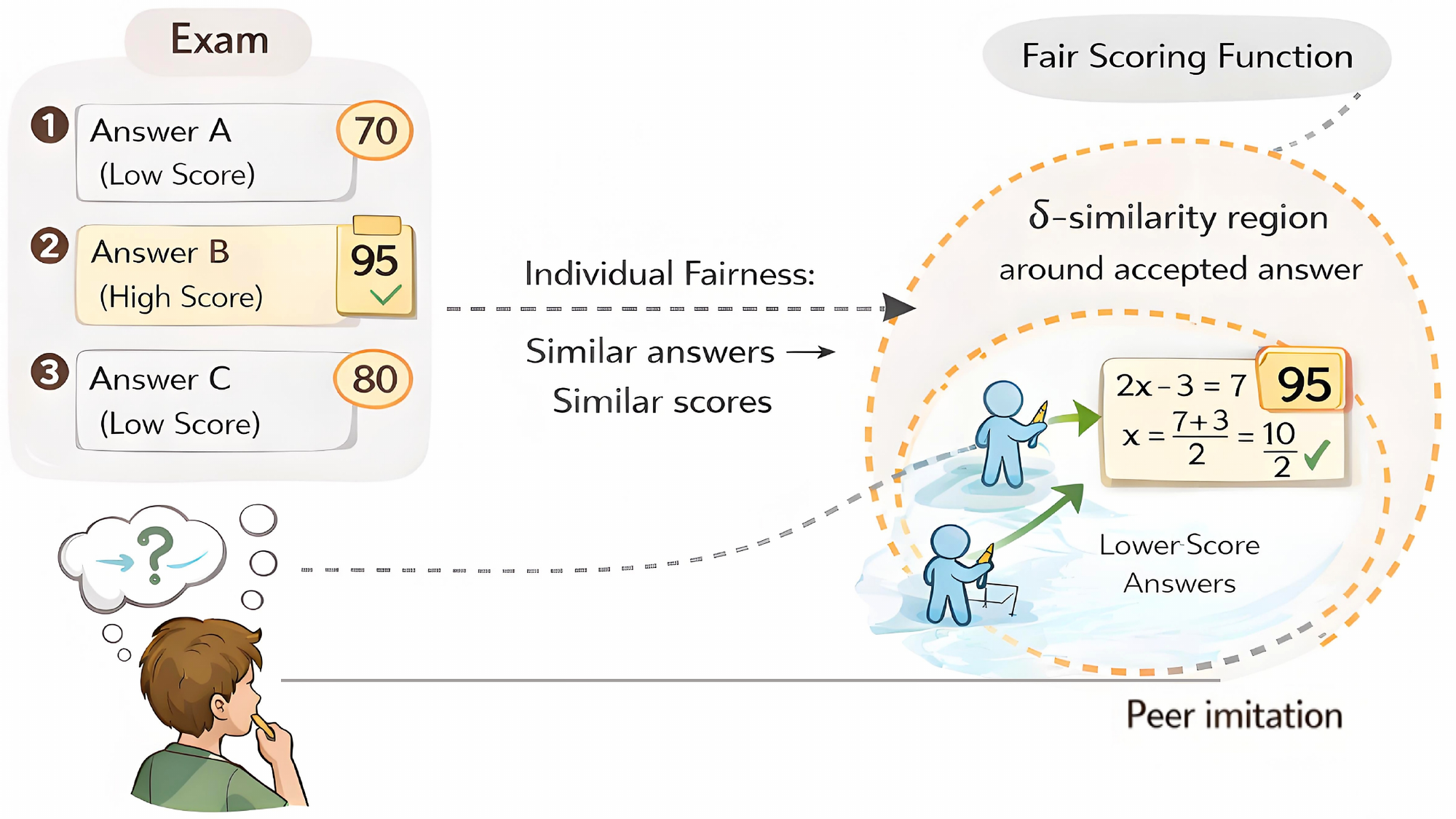}
        \caption{Examination scenario}
        \label{fige1a}
    \end{subfigure}
    \begin{subfigure}[t]{0.33\textwidth}
        \centering
        \includegraphics[width=\linewidth]{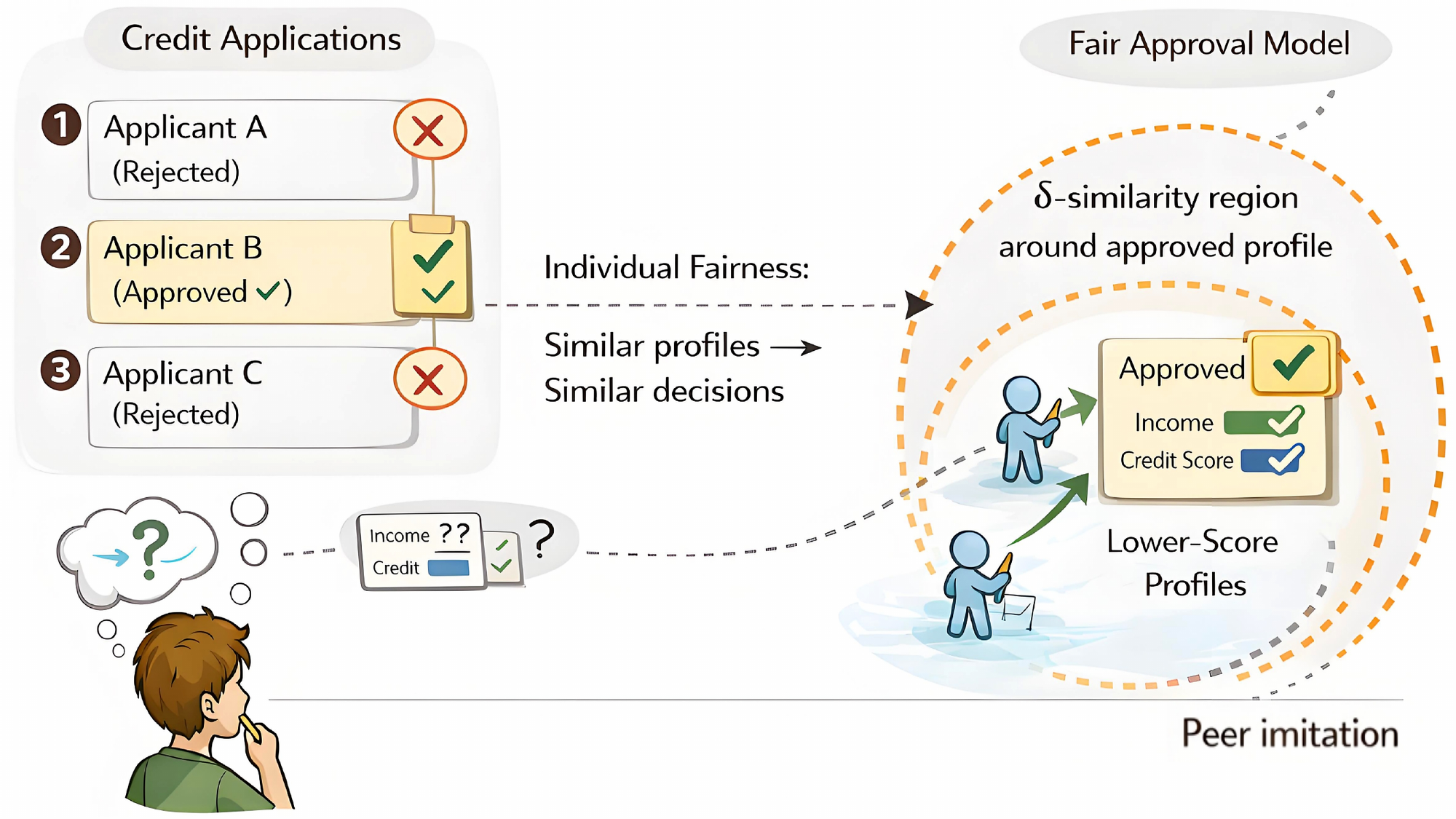}
        \caption{Credit approval scenario}
        \label{fige1b}
    \end{subfigure}
    \begin{subfigure}[t]{0.33\textwidth}
        \centering
        \includegraphics[width=\linewidth]{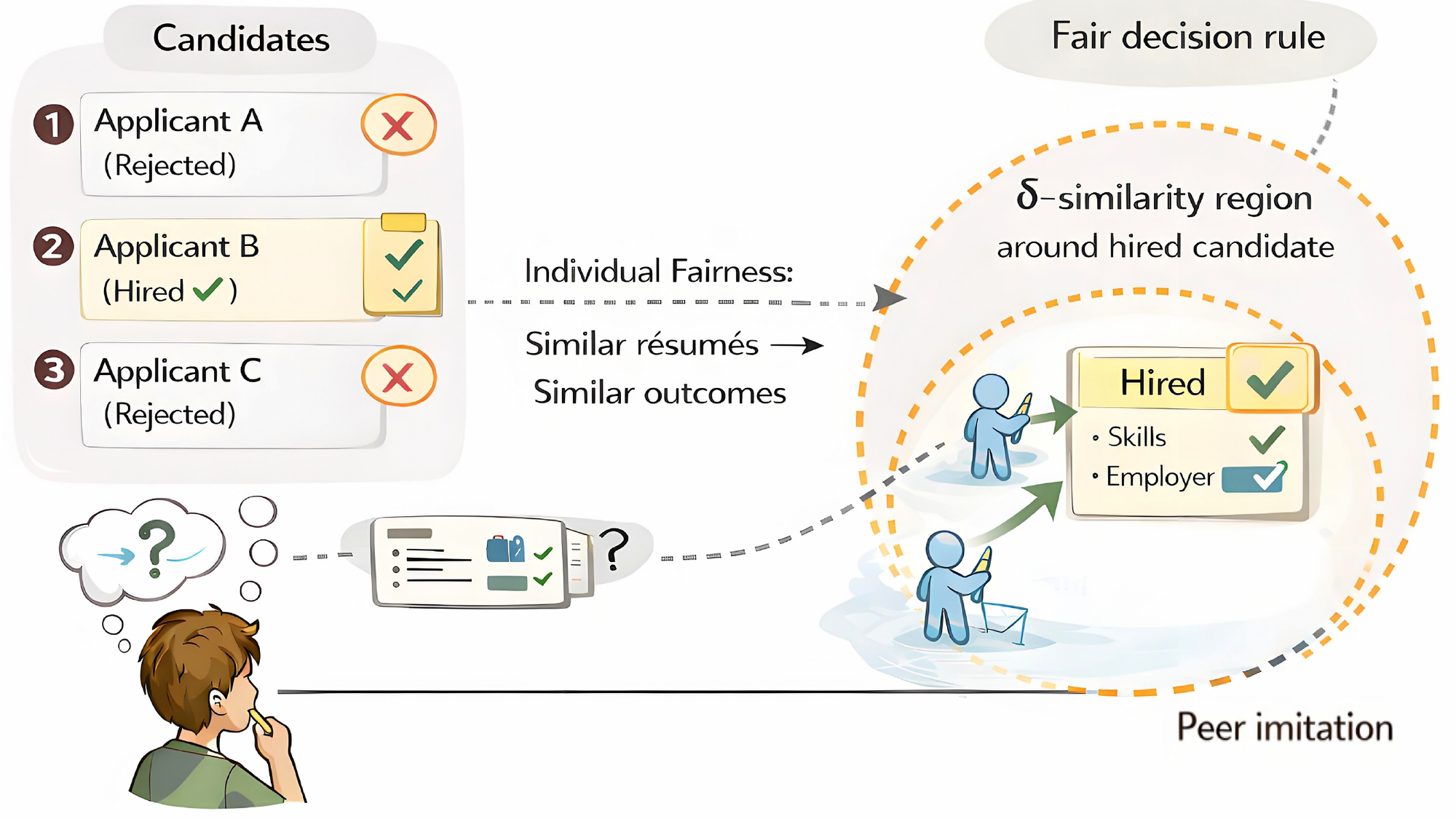}
        \caption{Hiring scenario}
        \label{fige1c}
    \end{subfigure}
    \vskip -0.08in
    \caption{Three examples of individual fairness in strategic classification. Under individual fairness, similar individuals are expected to receive consistent decisions, which naturally induces agents to adjust their features toward previously accepted peers. The examination, credit approval, and hiring scenarios illustrate this common behavior across domains.}
    \label{fige1}
    \vskip -0.1in
\end{figure*}

\subsection{Necessity of Individual Fairness for SC}
\label{sec:if_needed}

{\bf Intrinsic Requirement of Individual Fairness.} In strategic classification, individuals are commonly restricted to a local visibility, i.e., they tend to pay attention to several similar individuals, rather than disparate groups. Therefore, compared to group fairness (a global view in the group-level), the local view by individual fairness naturally arises at the level of~(neighboring) individuals, as similar individuals are expected to face similar opportunities for obtaining favorable outcomes under strategic adaptation. To further validate this viewpoint, we list three typical SC examples below:

\vskip 0.05in
\hspace{-8.3pt}
\shadowbox{\parbox{\dimexpr\linewidth-3\fboxsep-10\fboxrule\relax}{ \textbf{Examination~\cite{sacerdote2001peer,wesolowsky2000detecting}} As shown in Figure~\ref{fige1a}, in exams, students may adjust their answering strategies when uncertain about passing. Under individual fairness, similar answer patterns should receive similar scores. Students, therefore, adapt their responses to resemble those of higher-performing peers, as similarity to well-evaluated answers increases the chance of consistent grading. }}
\vskip 0.03in
\hspace{-8.3pt}
\shadowbox{\parbox{\dimexpr\linewidth-3\fboxsep-10\fboxrule\relax}{ \textbf{Credit approval~\cite{bjorkegren2020behavior,keswani2023addressing}} As shown in Figure~\ref{fige1b}, in credit approval, applicants may modify observable financial attributes to improve outcomes. With individual fairness, similar profiles are expected to receive consistent decisions. Once a certain profile is approved, nearby applicants may adjust their features to resemble it, moving into a region associated with positive decisions.}}
\vskip 0.03in
\hspace{-8.3pt}
\shadowbox{\parbox{\dimexpr\linewidth-3\fboxsep-10\fboxrule\relax}{ \textbf{Hiring~\cite{henle2019assessing,alder2006achieving}} As shown in Figure~\ref{fige1c}, in hiring, candidates may adapt credentials such as r\'{e}sum\'{e} structure or skill presentation. Under individual fairness, similar qualifications should lead to consistent decisions. Candidates, therefore, shape their profiles toward those of previously accepted applicants to fall within a favorable similarity region.}}
\vskip 0.01in

\noindent \textbf{Group fairness cannot Imply Individual Fairness in SC.}
Group fairness in Eq.~\eqref{eq:group_fair_sc} constrains only \emph{group-conditioned marginal} outcome rates under the post-manipulation distribution. In concrete, consider a fixed group $g$ and suppose there exist two post-manipulation feature points $\mathbf{u},\mathbf{v}$ with non-zero probability under the induced distribution such that
\begin{equation}
\begin{aligned}
    & d(\mathbf{u},\mathbf{v}) \le \delta, \\
    \text{with} &\qquad f(\mathbf{u};g) = 1, \quad f(\mathbf{v};g) = 0.
\end{aligned}
\label{eq:gf_not_imply_if_witness}
\end{equation}
The pair $(\mathbf{u},\mathbf{v})$ violates the individual fairness requirement in~\eqref{eq:if_sc_post}, since similar individuals receive different decisions.
However, the group-level statistic $\Pr(f(\mathbf{x}')=1\mid g)$ appearing in Eq.~\eqref{eq:group_fair_sc} depends only on the overall proportion of positive decisions within the group and is unaffected by how these decisions are distributed among similar individuals.
Thus, in the regime of SC tasks, satisfying group fairness does not imply individual fairness, as shown in the following proposition with proof in Appendix~\ref{appB}.

\begin{proposition}[Group fairness does not imply individual fairness]
\label{prop:gf_not_imply_if}
There exist a distribution $\mathcal{D}$, a classifier $f_g$ satisfying the group fairness constraint in Eq.~\eqref{eq:group_fair_sc}, and two post-manipulation feature points $\mathbf{u},\mathbf{v}$ such that
\begin{equation}
d(\mathbf{u},\mathbf{v}) \le \delta,
\qquad
f_g(\mathbf{u}) \neq f_g(\mathbf{v}),
\end{equation}
i.e., individual fairness in Eq.~\eqref{eq:if_sc_post} is violated despite satisfying group fairness.
\end{proposition}

\subsection{Peer Imitation of Individual Fairness Confounds Classical Strategic Classification}
\label{sec:ifsc_breakdown}

To further formulate the above intuitions and examples, we are aware of the fact that within the individual fairness constraint, an agent may instead manipulate by \emph{imitating} a nearby positively decided peer, aiming to enter its similarity neighborhood. This phenomenon, where we named ``Peer Imitation'', contrasts from the regime of typical paradigm of classical strategic classification, which assumes that agents respond \emph{independently} to the deployed classifier, i.e., each agent's strategic manipulation can be written as a function $b(x_i;f)$ (Eq.~\eqref{eq:best_response}):

\begin{definition}[Peer imitation as a feasibility constraint]
\label{def:peer_imitation}
Given a decision rule $f$ and a similarity metric $d$ with radius $\delta$, agent $i$ exhibits \emph{peer imitation} if its post-manipulation feature $\mathbf{x}_i'$ lies in the $\delta$-neighborhood of some positively decided individual $\{\mathbf{x}_j\}_{j=1}^m$, i.e.,
\begin{equation}
\begin{aligned}
\mathbf{x}'_i= & b_{IF}(\mathbf{x}_i; f_d, \{\mathbf{x}_j\}_{j=1}^m)
& \text{s.t.}\ \quad d(\mathbf{x}_i',\mathbf{x}_j)\le \delta\quad   \quad  f(\mathbf{x}_j)=1.
\end{aligned}
\label{eq:peer_imitation_def}
\end{equation}
\end{definition}

\noindent The key implication of Definition~\ref{def:peer_imitation} is that an agent's feasible manipulations depend on the location of \emph{positively decided peers}, i.e., $\mathbf{x}'_i = b_{\mathrm{IF}}\!\left(\mathbf{x}_i; f, \mathbf{x}_j\right)$, rather than on $(\mathbf{x}_i,f)$ alone. This peer imitation violates the independent assumption underlying classical strategic classification, which models the strategic manipulation in an agent-independent function $b(\mathbf{x}_i;f)$. Furthermore, we formalize this mismatch by the following proposition, with proof deferred to Appendix~\ref{app:proof_pro1}.

\begin{proposition}[Peer imitation is not captured by classical SC]
\label{pro1}
Peer imitation induces population-dependent manipulation: an agent's feasible post-manipulation feature depends on the locations of positively decided peers, which cannot be represented by an independent strategic manipulation.
In particular,
\begin{equation}
b_{\mathrm{IF}}\!\left(\mathbf{x}_i; f, \{\mathbf{x}_j\}_{j=1}^n\right)\ \not\equiv\ b(\mathbf{x}_i; f),
\label{eq:peer_not_separable}
\end{equation}
where $b_{\mathrm{IF}}$ denotes a peer-driven manipulation satisfying Eq.~\eqref{eq:peer_imitation_def} and $b(\cdot;f)$ is any independent manipulation function.
\end{proposition}

\section{Individual Fairness-aware Strategic Classification}
\label{sec:methods}

\subsection{Problem Re-Formulation}

Based on the emergence of the peer imitation phenomenon, we argue that the problem of classical SC tasks requires re-formulation by considering the interactions of positively decided peers:

\begin{problem}[Strategic Classification with Individual Fairness]
Given an individual fairness constraint $C_{if}$ specified by a similarity metric $d$ and radius $\delta$, a classifier $f$ is designed to achieve accurate predictions and maintain individually fair decisions under strategic manipulation via peer imitation.
\end{problem}

To address this problem, we propose \emph{IFSC}, a strategic framework that captures peer imitation under individual fairness and learns a classifier under the simulated strategic distribution.

\subsection{Strategic Manipulation via Visible Peers}
\label{sec:agent}
We model peer imitation under individual fairness with a visibility-driven manipulation mechanism.
Each agent first forms a local set of visible positively decided peers, and then performs peer imitation and moves into its individual-fairness neighborhood.

\noindent \textbf{Visible peers.}
We first define the set of positively decided peers visible to each agent.
Let $f$ denote the deployed classifier with individual fairness constraint and define the set $S^{+}(f)$ of positively decided agents as
\begin{equation}
S^{+}(f) = \{ j \mid f(\mathbf{x}_j)=1 \}.
\label{eq:positive_set}
\end{equation}

In practice, agents cannot observe all positively decided peers, but only those within a limited local neighborhood.
We therefore define the visible peer set $S_i^{\mathrm{vis}}(f)$ of a agent $i$ as follows:

\begin{definition}[Visible Peer Set]
Given a deployed classifier $f$ and the set of positively decided peers  \( S^{+}(f) \), the visible peer set $S_i^{\mathrm{vis}}(f)$ of agent $i$ is
\begin{equation}
S_i^{\mathrm{vis}}(f)
=
\{ j \in S^{+}(f) \mid k(\mathbf{x}_i,\mathbf{x}_j) \le R_{\mathrm{vis}} \},
\label{eq:visible_set}
\end{equation}
where $k(\cdot,\cdot)$ defines the neighborhood structure used for visibility of two agents and $R_{\mathrm{vis}}>0$ is the visibility radius.
\end{definition}
This reflects the realistic scenario that agents form strategic references from locally observable successful cases rather than the entire population.

\noindent \textbf{Peer-driven manipulation.}
Given a visible peer $j\in S_i^{\mathrm{vis}}(f)$, agent $i$ considers its utility for manipulating its features into the individual fairness neighborhood of $j$.
Therefore, each agent evaluates a strategic utility $U_i(\mathbf{x}'_i;j)$ to select the best peer for imitation.

\begin{definition}[Strategic Utility for Peer Imitation]
Given  a deployed classifier $f$ with individual-fairness similarity metric $d(\cdot,\cdot) \le\delta$, for each $j \in S_i^{\mathrm{vis}}(f)$, the strategic utility $U_i(\mathbf{x}'_i;j)$ for agent $i$ is defined as
\begin{equation}
\begin{aligned}
&U_i(\mathbf{x}^{(j)}_i;j)
=
f(\mathbf{x}^{(j)}_i) - c(\mathbf{x}_i,\mathbf{x}^{(j)}_i),
\quad \\
& \; \text{for } \; \mathbf{x}^{(j)}_i \;\text{ such that } \;\;d(\mathbf{x}^{(j)}_i,\mathbf{x}_j)\le\delta,
\end{aligned}
\label{eq:value_target}
\end{equation}
where $\mathbf{x}^{(j)}_i$ denote a candidate manipulation feature chosen by agent $i$ when imitating $j$ and $c(\mathbf{x}_i,\mathbf{x}'_i)$ denotes the manipulation cost.
\end{definition}

Besides, if no successful individual is visible, i.e., $S_i^{\mathrm{vis}}(f)=\emptyset$, the agent adopts a conservative fallback and does not perform peer imitation, setting $\mathbf{x}'_i=\mathbf{x}_i$.

Therefore, we characterize the strategic manipulation via peer imitation of agents, i.e., peer-driven manipulation, as follows:

\begin{definition}[Peer-driven Manipulation]
The strategic manipulation of agent $i$ under classifier $f$ with individual fairness is defined as
\begin{equation}
\begin{aligned}
\mathbf{x}'_i
&= b_{\mathrm{IF}}
\big(
\mathbf{x}_i;\,
f,\,
S_i^{\mathrm{vis}}(f)
\big) \\
&=
\begin{cases}
\displaystyle
\arg\max_{\substack{j \in S_i^{\mathrm{vis}}(f),\\
\mathbf{x}'_i : d(\mathbf{x}'_i,\mathbf{x}_{j})\le\delta}}
U_i(\mathbf{x}'_i;j),
& \text{if } S_i^{\mathrm{vis}}(f)\neq\emptyset,\\[7pt]
\mathbf{x}_i,
& \text{otherwise}.
\end{cases}
\end{aligned}
\label{eq:if_manipulation}
\end{equation}
\end{definition}

\subsection{Robust Learning against Peer-driven Manipulation}
\label{sec:vr_training}

Under individual fairness, peer-driven manipulation is conditioned on which positively decided peers are visible to each agent, i.e., the visibility set $S_i^{\mathrm{vis}}(f)$.
In practice, this visibility is uncertain: an agent may observe only a subset of nearby positive peers, or the visible set can vary across environments.

To account for this uncertainty, we adopt a \emph{peer imitation-robust learning} procedure, in which stochastic perturbations of visible peer sets are used to simulate realized peer-driven manipulations during training.

\noindent \textbf{Peer-driven manipulation outcome simulation.}
We simulate peer imitation outcomes under perturbed visible peer sets rather than assuming a fixed visibility realization.

Specifically, during training the visible peer set $S_i^{\mathrm{vis}}(f)$ is randomly perturbed for construct a perturbed peer set $\widetilde{S}_i^{\mathrm{vis}}(f)$, i.e., $\widetilde{S}_i^{\mathrm{vis}}(f) \sim \mathcal{T}\big(S_i^{\mathrm{vis}}(f)\big)$, where $\mathcal{T}(\cdot)$ denotes a stochastic perturbation operator that randomly removes a subset of visible peers and introduces a small number of alternative positively decided peers.

Therefore, the resulting simulated post-manipulation features for training are defined as
\begin{equation}
\widetilde{\mathbf{x}}'_i
=
b_{\mathrm{IF}}
\big(
\mathbf{x}_i;\,
f,\,
\widetilde{S}_i^{\mathrm{vis}}(f)
\big).
\label{eq:vr_manipulation}
\end{equation}

The manipulation rule remains identical to Eq.~\eqref{eq:if_manipulation}, but is evaluated under the perturbed visibility.

\noindent \textbf{Classifier learning against peer-driven manipulation.}
Given the simulated peer-driven manipulation distribution, the classifier is trained with the following optimization objective:
\begin{definition}[Classifier Learning Objective]
Let $\mathcal{L}(\cdot,\cdot)$ denote a classification loss. The robust learning objective is defined as
\begin{equation}
\min\mathcal{R}_{\mathrm{acc}}(f)
\;=\; \min_{f}
 \mathbf{E}_{(\mathbf{x},y)\sim\mathcal{D}}
\Big[
\mathcal{L}\big(
f(\widetilde{\mathbf{x}}'),y
\big)
\Big],
\label{eq:post_risk}
\end{equation}
where $\widetilde{\mathbf{x}}'$ is the simulated peer-driven manipulation features.
\end{definition}

This procedure prevents the classifier from overfitting to a specific visibility realization and improves robustness under individual-fairness-aware strategic manipulation.

\subsection{Unified IFSC Optimization}
\label{sec:framework}

We now incorporate an individual fairness regularizer and present the unified IFSC optimization objective.

\noindent \textbf{Individual fairness regularization.}

Since decisions at deployment are made on post-manipulation inputs, individual fairness is enforced on the same induced feature space.
We define the following individual fairness regularization:
\begin{equation}
\mathcal{V}_{\mathrm{IF}}(f)
=
\frac{1}{|\mathcal{P}_\delta|}
\sum_{(i,j)\in\mathcal{P}_\delta}
\big|
f(\widetilde{\mathbf{x}}'_i)-f(\mathbf{x}_j)
\big|,
\label{eq:if_violation}
\end{equation}
where $\mathcal{P}_\delta = \big\{(i,j)\mid i\neq j, d(\widetilde{\mathbf{x}}'_i,\mathbf{x}_j)\le\delta\big\}$ denotes the $\delta$-neighbor pairs, $d(\cdot,\cdot)$ is the individual-fairness similarity metric, and $\delta>0$ is the fairness threshold.

\noindent \textbf{Unified IFSC objective.}
We now combine the post-manipulation accuracy risk and the individual-fairness penalty into a single optimization objective.
Let $\mathcal{F}$ denote the hypothesis class and let $\lambda>0$ control the accuracy-fairness trade-off.
The IFSC classifier is learned by solving
\begin{equation}
f^\star_{IF}
\in
\arg\min_{f\in\mathcal{F}}
\Big[
\mathcal{R}_{\mathrm{acc}}(f)
+
\lambda\,\mathcal{V}_{\mathrm{IF}}(f)
\Big],
\label{eq:ifsc_outer}
\end{equation}
where $\mathcal{R}_{\mathrm{acc}}(f)$ is defined in~\eqref{eq:post_risk} and $\mathcal{V}_{\mathrm{IF}}(f)$ is defined in~\eqref{eq:if_violation}.
Notably, both terms are evaluated on the simulated post-manipulation features $\widetilde{\mathbf{x}}'$ induced by Eq.~\eqref{eq:vr_manipulation}.

The whole process of the individual fairness-aware strategic classification framework is illustrated in Algorithm~\ref{alg:ifsc}.

\section{Experiment}
We evaluate our IFSC framework from the following questions:
\begin{itemize}[leftmargin=18pt]
    \item Whether peer-driven manipulation is more effective than independent manipulation.
    \item Whether IFSC improves post-manipulation accuracy and individual fairness across manipulation regimes.
    \item Whether IFSC remains robust under visibility mismatch and stochastic perturbation.
\end{itemize}

\begin{algorithm}[t]
\centering
\caption{Individual Fairness-Aware Strategic Classification (IFSC)}
\label{alg:ifsc}
\begin{algorithmic}[1]
\Require Training data $\mathcal{D}=\{(x_i,y_i)\}_{i=1}^n$; A classifier $f \in \mathcal{F}$ with individual fairness constraint; fairness weight $\gamma$
\State Build visibility set $S_i^{\text{vis}}(f)$ of each agent $i$ with $\mathcal{V}_i$~(Eq.~\eqref{eq:visible_set})
\State Simulate peer-driven manipulation $\widetilde{\mathbf{x}}'_i$ (Eq.~\eqref{eq:vr_manipulation})
\State Form post-manipulation data $\mathcal{D}'=\{(\widetilde{\mathbf{x}}'_i,y_i)\}_{i=1}^n$
\State Train strategic classifier $f^\star_{IF}$ with individual fairness constraint on $\mathcal{D}'$~(Eq.~\eqref{eq:ifsc_outer})
\State Return $f^\star_{IF}$
\end{algorithmic}
\end{algorithm}

\subsection{Experimental Setup}

\textbf{Datasets.}
We evaluate our framework on seven datasets, including six real-world benchmarks and one synthetic dataset:
\begin{itemize}[leftmargin=18pt]
    \item \textbf{Credit}~\cite{default_of_credit_card_clients_350}: Credit card default prediction based on financial attributes.
    \item \textbf{Adult}~\cite{adult_2}: Income classification from U.S.\ census data.
    \item \textbf{Diabetes}~\cite{Teboul2015Diabetes}: Diabetes risk prediction using clinical and demographic features.
    \item \textbf{German}~\cite{statlog_(german_credit_data)_144}: Credit risk classification based on personal and financial profiles.
    \item \textbf{Spam}~\cite{spambase_94}: Email spam detection using textual and statistical features.
    \item \textbf{PhiUSIIL}~\cite{phiusiil_phishing_url_(website)_967}: Phishing URL detection based on lexical and structural URL features.
    \item \textbf{Synthetic}~\cite{lopez2016paysim}: Simulated mobile transaction data for fraud detection.
\end{itemize}

\noindent \textbf{Baselines.}
All methods employ linear classifiers, consistent with standard practice in strategic classification~\cite{chen2023learning,shavit2020causal,pmlr-v139-ghalme21a}.
We compare two strategic classifiers that share the \emph{same linear model class}, manipulation cost, and individual-fairness regularization, and differ only in the \emph{assumed agent manipulation model} used during training.
\begin{itemize}[leftmargin=18pt]
    \item \textit{Standard SC} follows classical strategic classification. During training, agents are assumed to respond via \emph{independent} manipulation (Eq.~\eqref{eq:best_response}). Individual fairness is incorporated at the classifier level through an output regularization term.
    \item Our proposed method (\textbf{IFSC}) uses the same linear classifier, cost structure, and individual fairness constraint, but replaces independent manipulation with \emph{peer-driven manipulation}.
\end{itemize}

\noindent \textbf{Manipulation regimes.}
We evaluate classifiers under three strategic manipulation regimes.
\begin{itemize}[leftmargin=18pt]
    \item \textbf{Classical manipulation}, where agents independently modify features via best-response optimization under a Mahalanobis cost, as in standard strategic classification.
    \item \textbf{Peer imitation}, where agents imitate nearby positively classified individuals by moving into their neighborhoods, inducing similarity-dependent and interdependent responses.
    \item \textbf{Mixed manipulation}, where agents follow classical best-response or individual-fairness-aware imitation with equal probability.
\end{itemize}

\noindent \textbf{Metrics.}
We evaluate classifiers on the post-manipulation distribution using accuracy and an individual fairness metric.
\begin{itemize}[leftmargin=18pt]
    \item \textbf{Accuracy (Acc)}, measuring predictive performance under strategic manipulation,
    \begin{equation}
    \mathrm{Acc}
    =
    \frac{1}{n}
    \sum_{i=1}^{n}
    \mathbb{I}\!\left[ f(\mathbf{x}'_i) = y_i \right].
    \end{equation}

    \item \textbf{Individual Fairness Gap (IF-Gap)}, which quantifies decision inconsistency among similar individuals after manipulation~\cite{lahoti2019ifair,gupta2021individual}.
    Let $d(\cdot,\cdot)$ denote the individual fairness similarity metric and $\delta>0$ a similarity threshold.
    We define
    \begin{equation}
    \mathrm{IF\text{-}Gap}
    =
    \mathbb{E}_{(i,j)}
    \big[
    \bbold{1}\!\left[
    f(\mathbf{x}'_i)\neq f(\mathbf{x}_j)
    \right]
    \;\big|\;
    d(\mathbf{x}'_i,\mathbf{x}_j)\le\delta
    \big],
    \end{equation}
    where the expectation is taken over all $\delta$-neighbor pairs. Lower values indicate stronger individual fairness.
    \item \textbf{Cheatment rate.} To quantify the effectiveness of strategic manipulation, the \emph{cheatment rate} captures how often individuals who are initially rejected can become accepted after manipulation. Let $\mathcal{N}=\{i \mid f_{\mathrm{IF}}(x_i)=0\}$ denote the set of initially rejected instances. Given manipulation mechanism $m$ and post-manipulation features $x'_i=b_m(x_i;f_{\mathrm{IF}})$, we define
    \begin{equation}
    R_{cheat}
    =
    \frac{1}{|\mathcal{N}|}
    \sum_{i\in\mathcal{N}}
    \bbold{1}\!\left[f_{\mathrm{IF}}(x'_i)=1\right].
    \end{equation}
    A higher rate indicates that the deployed classifier is more susceptible to the corresponding manipulation.
\end{itemize}

\begin{figure*}[t]
    \centering
    \begin{subfigure}[b]{0.239\textwidth}
        \includegraphics[width=\linewidth]{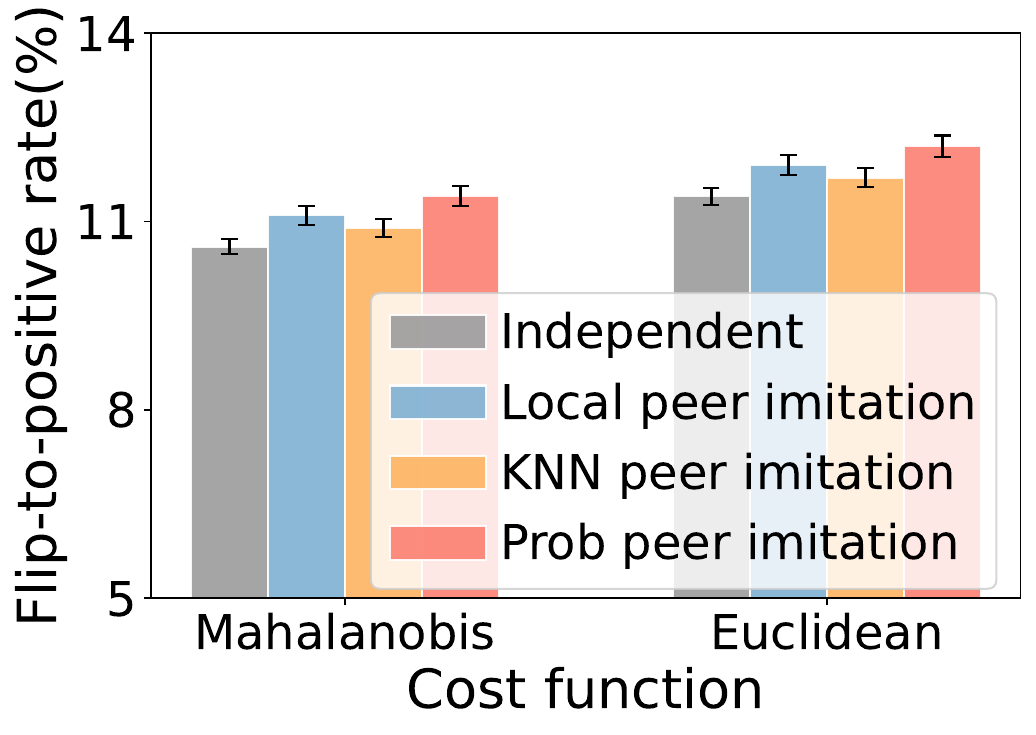}
        \caption{$d(\mathbf{x}'_i, \mathbf{x}_j) \le 4 $}
        \label{fig2a}
    \end{subfigure}
    \begin{subfigure}[b]{0.239\textwidth}
        \includegraphics[width=\linewidth]{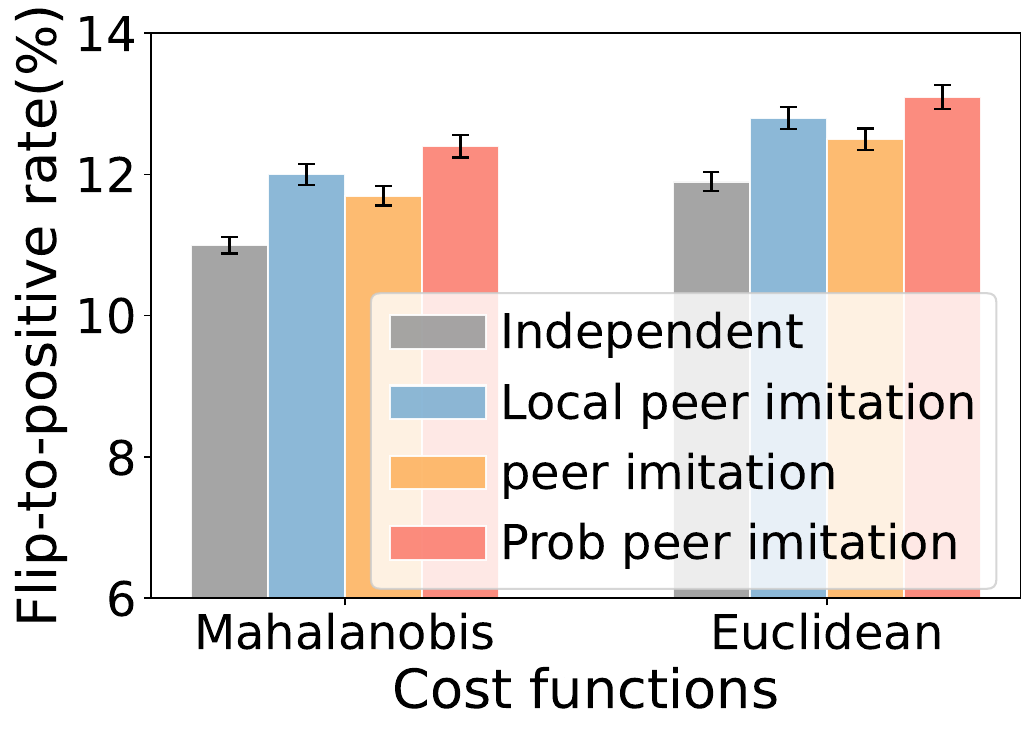}
        \caption{$d(\mathbf{x}'_i, \mathbf{x}_j) \le 5.2 $}
        \label{fig2b}
    \end{subfigure}
    \begin{subfigure}[b]{0.239\textwidth}
        \includegraphics[width=\linewidth]{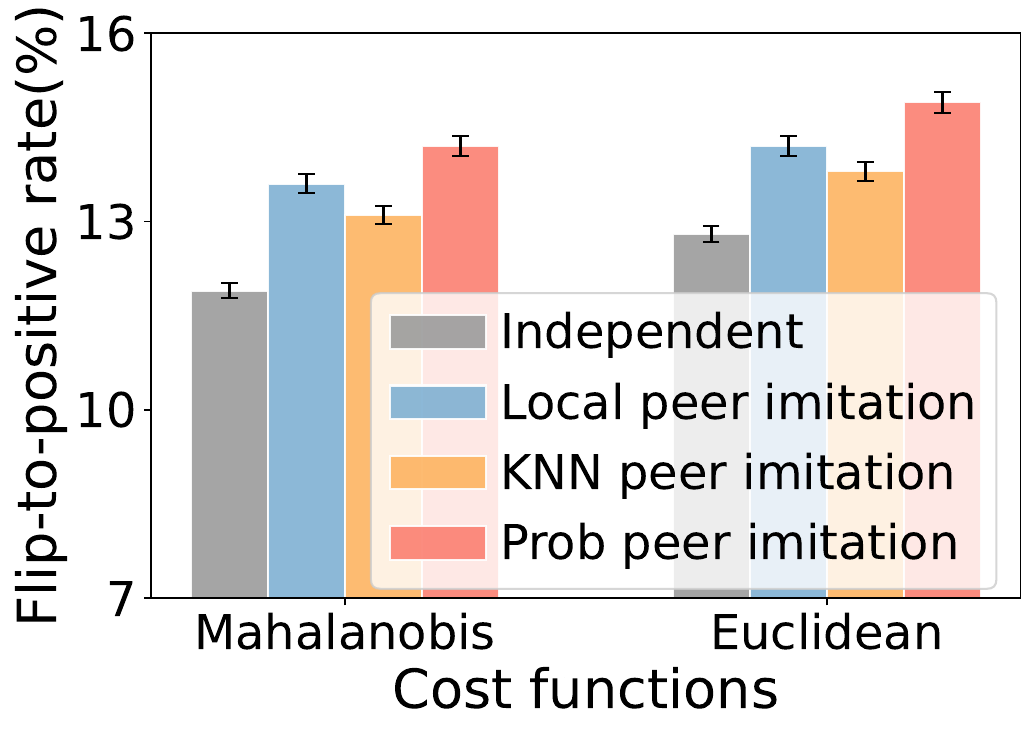}
        \caption{$d(\mathbf{x}'_i, \mathbf{x}_j) \le 5.9 $}
        \label{fig2c}
    \end{subfigure}
    \begin{subfigure}[b]{0.239\textwidth}
        \includegraphics[width=\linewidth]{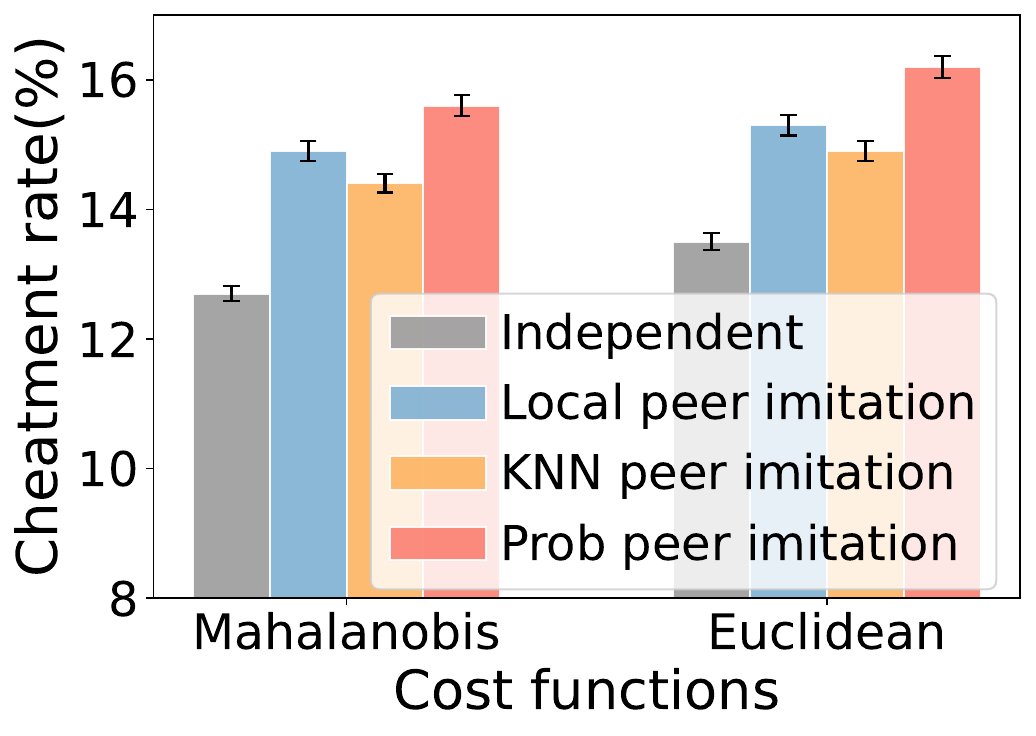}
        \caption{$d(\mathbf{x}'_i, \mathbf{x}_j) \le 6.8 $}
        \label{fig2d}
    \end{subfigure}
    \vskip -0.1in
    \caption{Cheatment rate under different manipulation costs. Performances compare independent manipulation and three Peer imitation~(ours) manipulations (local, kNN-based, probabilistic visibility) under Mahalanobis and Euclidean costs. Subfigures (a)--(d) correspond to increasing individual fairness thresholds $\delta= [4,5.2,5.9,6.8]$, calibrated to the $5\%$, $10\%$, $15\%$, and $20\%$ of standardized Euclidean distances.}
    \label{fig2}
    \vskip -0.05in
\end{figure*}

\begin{table*}[t]
\centering
\vskip -0.03in
\caption{Post-manipulation accuracy (\%) under different manipulation regimes.}
\vskip -0.1in
\label{tab:acc_main}
\resizebox{\textwidth}{!}{
\begin{threeparttable}
\small
\setlength{\tabcolsep}{3.6pt}
\begin{tabular}{l l c c c c c c c}
\toprule
\textbf{Manipulation} & \textbf{Classifier}
& \textit{Adult} & \textit{Credit} & \textit{Diabetes} & \textit{German}
& \textit{Spam} & \textit{PhiUSIIL} & \textit{Synthetic} \\
\midrule

\multirow{2}{*}{\textit{Classical}}
& \textbf{Standard SC}
& $\mathbf{79.62_{\pm0.91}}$ & $77.48_{\pm1.22}$ & $71.36_{\pm1.41}$
& $75.10_{\pm1.33}$ & $\mathbf{85.12_{\pm1.09}}$
& $\mathbf{83.04_{\pm1.18}}$ & $\mathbf{83.05_{\pm1.18}}$ \\
& \textbf{IFSC (ours)}
& $78.94_{\pm0.88}$ & $\mathbf{78.31_{\pm1.17}}$ & $\mathbf{72.02_{\pm1.32}}$
& $\mathbf{75.82_{\pm1.29}}$ & $84.76_{\pm1.12}$
& $82.88_{\pm1.21}$ & $82.64_{\pm1.10}$ \\
\midrule
\multirow{2}{*}{\textit{Peer imitation}}
& \textbf{Standard SC}
& $74.08_{\pm1.76}$ & $71.02_{\pm1.95}$ & $66.35_{\pm1.68}$
& $69.44_{\pm1.87}$ & $79.88_{\pm1.54}$
& $77.96_{\pm1.72}$ & $75.92_{\pm1.96}$ \\
& \textbf{IFSC (ours)}
& $\mathbf{83.22_{\pm1.14}}$ & $\mathbf{83.96_{\pm1.28}}$ & $\mathbf{75.38_{\pm1.19}}$
& $\mathbf{80.41_{\pm1.36}}$ & $\mathbf{90.62_{\pm1.05}}$
& $\mathbf{88.94_{\pm1.13}}$ & $\mathbf{86.14_{\pm1.22}}$ \\
\midrule
\multirow{2}{*}{\textit{Mixed}}
& \textbf{Standard SC}
& $76.12_{\pm1.39}$ & $73.46_{\pm1.61}$ & $68.42_{\pm1.57}$
& $71.28_{\pm1.65}$ & $81.06_{\pm1.42}$
& $79.14_{\pm1.59}$ & $78.26_{\pm1.71}$ \\
& \textbf{IFSC (ours)}
& $\mathbf{80.56_{\pm1.26}}$ & $\mathbf{80.92_{\pm1.41}}$ & $\mathbf{73.02_{\pm1.27}}$
& $\mathbf{77.54_{\pm1.48}}$ & $\mathbf{87.44_{\pm1.23}}$
& $\mathbf{85.76_{\pm1.31}}$ & $\mathbf{84.03_{\pm1.35}}$ \\
\midrule

\multirow{2}{*}{\textit{Noisy}}
& \textbf{Standard SC}
& $79.18_{\pm1.02}$ & $75.88_{\pm1.35}$ & $70.52_{\pm1.46}$
& $\mathbf{74.34_{\pm1.31}}$ & $84.93_{\pm1.11}$
& $82.21_{\pm1.23}$ & $82.34_{\pm1.28}$ \\
& \textbf{IFSC (ours)}
& $\mathbf{79.27_{\pm0.97}}$ & $\mathbf{76.42_{\pm1.22}}$ & $\mathbf{72.18_{\pm1.33}}$
& $74.05_{\pm1.37}$ & $\mathbf{85.31_{\pm1.08}}$
& $\mathbf{83.46_{\pm1.18}}$ & $\mathbf{83.17_{\pm1.21}}$ \\
\bottomrule
\end{tabular}
\begin{tablenotes}[para,flushleft]
\emph{Notes:} \textbf{Standard SC} follows classical strategic classification with linear classifiers, where agents respond via independent manipulation in Eq.~\eqref{eq:best_response}~\cite{hardt2016strategic,pmlr-v139-ghalme21a,shavit2020causal,chen2023learning}.
\textit{Noisy} denotes a distribution-mismatched manipulation setting that is not explicitly covered by either method during training.
\end{tablenotes}
\end{threeparttable}
}
\end{table*}

\begin{table*}[t]
\centering
\vskip -0.03in
\caption{Individual Fairness Gap (lower is better) under different manipulation regimes.}
\vskip -0.08in
\label{tab:ifgap_main}
\resizebox{\textwidth}{!}{
\begin{threeparttable}
\small
\setlength{\tabcolsep}{3.6pt}
\begin{tabular}{l l c c c c c c c}
\toprule
\textbf{Manipulation} & \textbf{Classifier}
& \textit{Adult} & \textit{Credit} & \textit{Diabetes} & \textit{German}
& \textit{Spam} & \textit{PhiUSIIL} & \textit{Synthetic} \\
\midrule

\multirow{2}{*}{\textit{Classical}}
& \textbf{Standard SC}
& $0.242_{\pm0.018}$ & $0.268_{\pm0.021}$ & $0.231_{\pm0.019}$
& $0.284_{\pm0.023}$ & $0.176_{\pm0.014}$
& $0.192_{\pm0.016}$ & $0.148_{\pm0.013}$ \\
& \textbf{IFSC (ours)}
& $\mathbf{0.198_{\pm0.016}}$ & $\mathbf{0.221_{\pm0.018}}$ & $\mathbf{0.186_{\pm0.017}}$
& $\mathbf{0.238_{\pm0.020}}$ & $\mathbf{0.149_{\pm0.013}}$
& $\mathbf{0.162_{\pm0.014}}$ & $\mathbf{0.118_{\pm0.010}}$ \\
\midrule

\multirow{2}{*}{\textit{Peer imitation}}
& \textbf{Standard SC}
& $0.184_{\pm0.016}$ & $0.206_{\pm0.018}$ & $0.172_{\pm0.017}$
& $0.221_{\pm0.020}$ & $0.132_{\pm0.012}$
& $0.148_{\pm0.014}$ & $0.102_{\pm0.010}$ \\
& \textbf{IFSC (ours)}
& $\mathbf{0.112_{\pm0.011}}$ & $\mathbf{0.128_{\pm0.013}}$ & $\mathbf{0.104_{\pm0.012}}$
& $\mathbf{0.142_{\pm0.015}}$ & $\mathbf{0.072_{\pm0.008}}$
& $\mathbf{0.084_{\pm0.009}}$ & $\mathbf{0.051_{\pm0.006}}$ \\
\midrule

\multirow{2}{*}{\textit{Mixed}}
& \textbf{Standard SC}
& $0.211_{\pm0.017}$ & $0.236_{\pm0.020}$ & $0.198_{\pm0.018}$
& $0.252_{\pm0.021}$ & $0.154_{\pm0.013}$
& $0.168_{\pm0.015}$ & $0.124_{\pm0.011}$ \\
& \textbf{IFSC (ours)}
& $\mathbf{0.146_{\pm0.013}}$ & $\mathbf{0.162_{\pm0.015}}$ & $\mathbf{0.138_{\pm0.014}}$
& $\mathbf{0.184_{\pm0.017}}$ & $\mathbf{0.098_{\pm0.010}}$
& $\mathbf{0.112_{\pm0.011}}$ & $\mathbf{0.074_{\pm0.008}}$ \\
\midrule

\multirow{2}{*}{\textit{Noisy}}
& \textbf{Standard SC}
& $0.232_{\pm0.019}$ & $0.255_{\pm0.022}$ & $0.219_{\pm0.020}$
& $0.271_{\pm0.024}$ & $0.168_{\pm0.015}$
& $0.181_{\pm0.017}$ & $0.139_{\pm0.014}$ \\
& \textbf{IFSC (ours)}
& $\mathbf{0.186_{\pm0.015}}$ & $\mathbf{0.209_{\pm0.017}}$ & $\mathbf{0.178_{\pm0.016}}$
& $\mathbf{0.226_{\pm0.019}}$ & $\mathbf{0.141_{\pm0.012}}$
& $\mathbf{0.153_{\pm0.013}}$ & $\mathbf{0.109_{\pm0.009}}$ \\

\bottomrule
\end{tabular}
\begin{tablenotes}[para,flushleft]
\emph{Notes:} Individual Fairness Gap is reported as a value in $[0,1]$, measuring the rate of decision inconsistency among $\delta$-similar individuals after manipulation.
\end{tablenotes}
\end{threeparttable}
}
\end{table*}

\begin{figure*}[t]
    \centering
    \begin{subfigure}[b]{0.239\textwidth}
        \includegraphics[width=\linewidth]{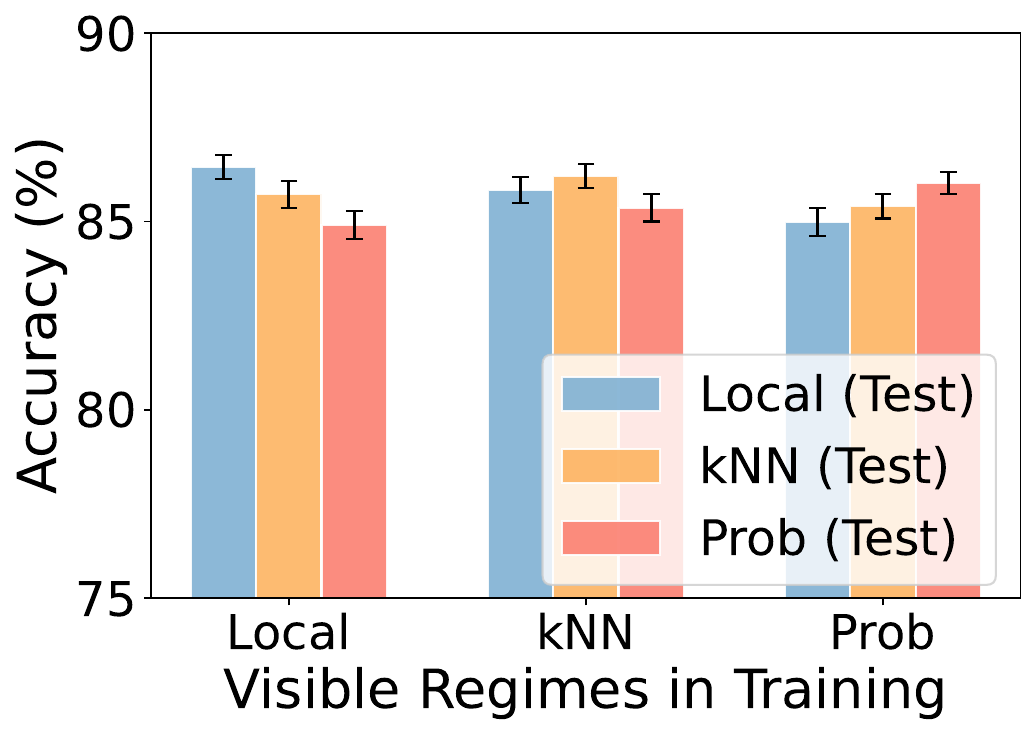}
        \caption{$d(\mathbf{x}'_i, \mathbf{x}_j) \le 4 $}
        \label{fig5a}
    \end{subfigure}
    \begin{subfigure}[b]{0.239\textwidth}
        \includegraphics[width=\linewidth]{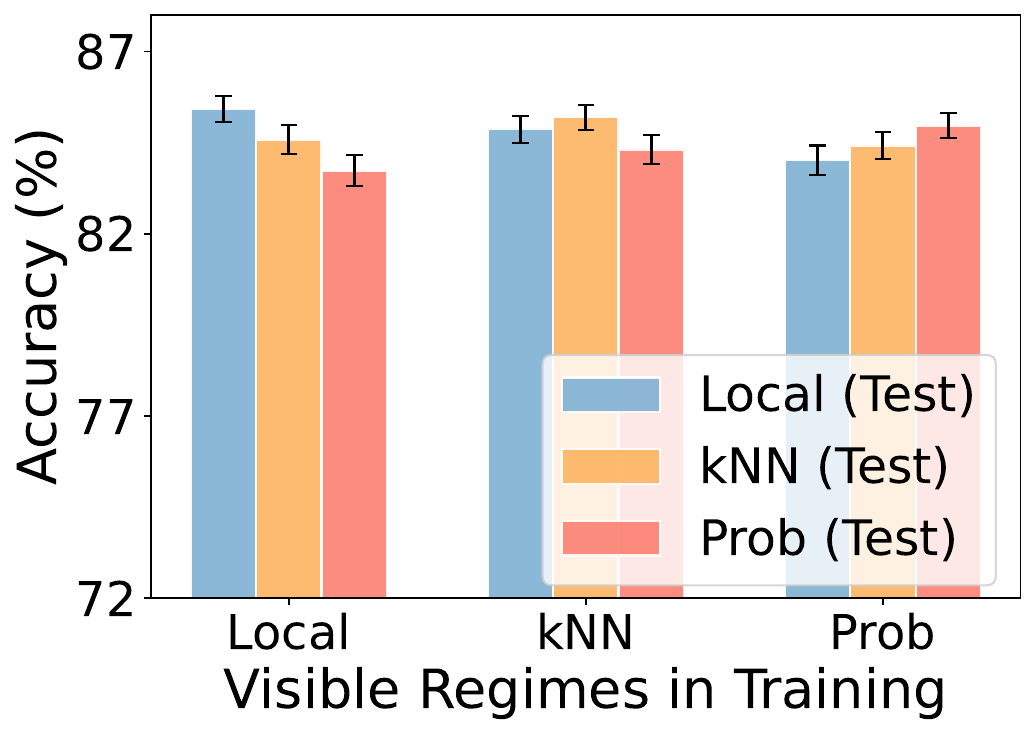}
        \caption{$d(\mathbf{x}'_i, \mathbf{x}_j) \le 6 $}
        \label{fig5b}
    \end{subfigure}
    \begin{subfigure}[b]{0.239\textwidth}
        \includegraphics[width=\linewidth]{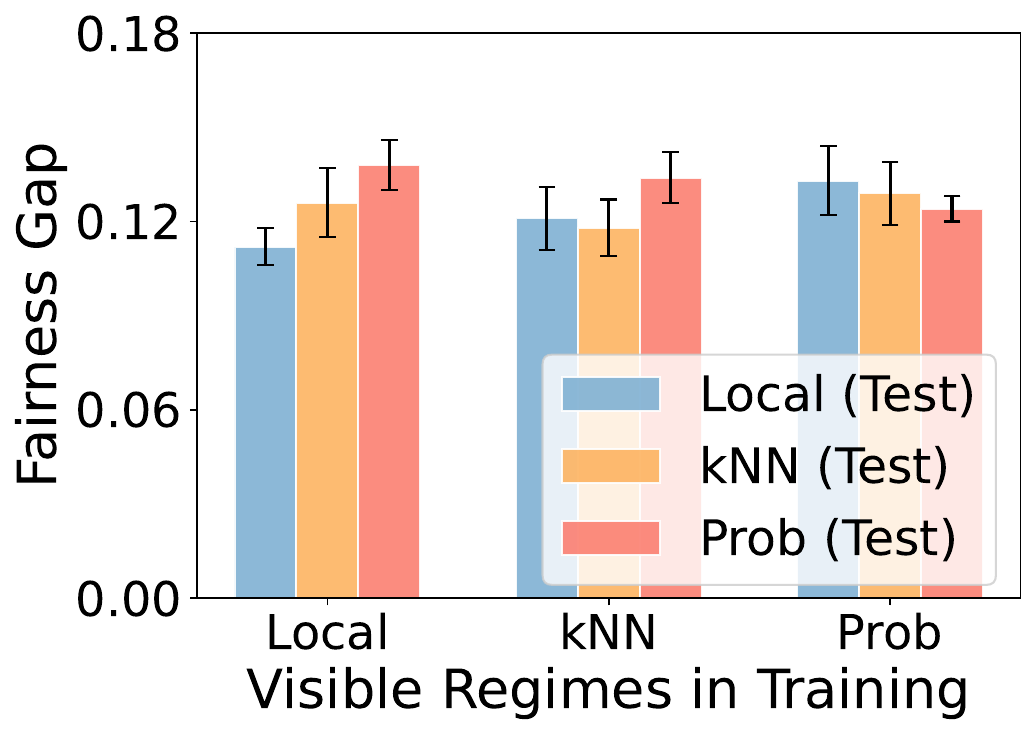}
        \caption{$d(\mathbf{x}'_i, \mathbf{x}_j) \le 4 $}
        \label{fig5c}
    \end{subfigure}
    \begin{subfigure}[b]{0.239\textwidth}
        \includegraphics[width=\linewidth]{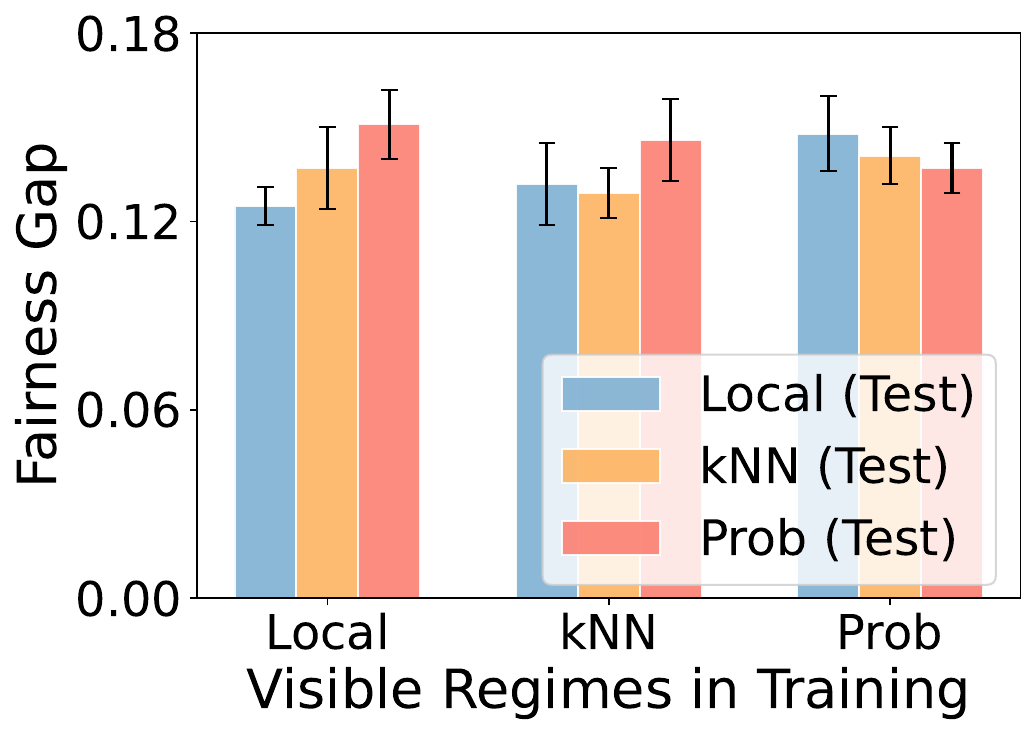}
        \caption{$d(\mathbf{x}'_i, \mathbf{x}_j) \le 6 $}
        \label{fig5d}
    \end{subfigure}
    \vskip -0.1in
    \caption{Post-manipulation accuracy (left two panels) and individual fairness gap (right two panels) under different visibility regimes. $d(\cdot,\cdot)$ is the individual similarity metric for individual fairness. The x-axis denotes the visibility regime used during training, and bars indicate performance under different test-time visibility regimes.}
    \label{fig5}
    \vskip -0.1in
\end{figure*}

\subsection{Visibility Regimes under Individual Fairness}
\label{sec:visibility}

The visibility set $S_i^{\mathrm{vis}}(f)$ specifies which positively decided peers agent $i$ can observe and use as strategic references under the deployed classifier $f$.
It determines the information available to agents when forming peer-imitation strategies, and is conceptually independent of the manipulation cost and the classifier optimization objective.
In practice, agents may observe successful peers through different information channels, which are often local, partial, and uncertain.
We therefore consider three representative visibility regimes.

\noindent \textbf{Metric-based local visibility.}
The most direct assumption is that agents observe nearby successful peers under the individual-fairness metric.
Given a visibility radius $R_{\mathrm{vis}}>0$, we define
\begin{equation}
S_i^{\mathrm{vis}}(f)
=
\{j \in S^{+}(f) \mid d(\mathbf{x}_i,\mathbf{x}_j) \le R_{\mathrm{vis}}\}.
\end{equation}
This regime captures deterministic local observability, where agents mainly compare themselves with successful individuals who are similar under the same metric used for individual fairness.

\noindent \textbf{Salient-feature kNN visibility.}
Agents may also form comparisons using only a subset of observable or salient attributes rather than the full feature representation.
Let $\mathcal{A}\subseteq\{1,\ldots,d\}$ denote the salient feature subset and let $\mathbf{x}^{(\mathcal{A})}$ be the corresponding projection.
The visible peer set is defined as
\begin{equation}
S_i^{\mathrm{vis}}(f)
=
\mathrm{kNN}_k
\Big(
\mathbf{x}_i^{(\mathcal{A})};
\{\mathbf{x}_j^{(\mathcal{A})}:j\in S^{+}(f)\}
\Big).
\end{equation}
This regime models partial observability, where agents imitate accepted peers based on limited or socially salient information.

\noindent \textbf{Probabilistic visibility kernel.}
Finally, visibility can be stochastic due to incomplete information or heterogeneous exposure.
For each positively decided peer $j\in S^{+}(f)$, we define the observation probability as
\begin{equation}
\Pr\!\left(j\in S_i^{\mathrm{vis}}(f)\right)
=
\rho \exp\!\left(
-\frac{d(\mathbf{x}_i,\mathbf{x}_j)^2}{\sigma^2}
\right),
\end{equation}
where $\sigma>0$ controls the spatial range of visibility and $\rho\in(0,1]$ controls the overall observation rate.
A smaller $\sigma$ makes visibility more local, while a larger $\sigma$ allows agents to observe more distant successful peers.

Together, these regimes cover hard local visibility, partial-feature visibility, and stochastic observability.
They are used to instantiate $S_i^{\mathrm{vis}}(f)$ in the peer-driven manipulation simulation and to evaluate the robustness of IFSC under different visibility assumptions.

\begin{remark}
These visibility regimes reflect three common forms of information access in real-world strategic settings.
Metric-based local visibility corresponds to high-transparency cases, where agents can observe relatively complete profiles of similar successful peers, such as accepted job or loan applicants.
Salient-feature kNN visibility captures partial observability, where agents only compare a few salient attributes, such as exam patterns or basic clinical features.
Probabilistic visibility models noisy exposure, where successful cases are learned indirectly through word-of-mouth, online forums, social media, or recommendation systems.
Thus, the three regimes respectively represent complete local information, partial salient information, and stochastic information access.
\end{remark}

\subsection{Implementation Details}
\label{subsec:implementation}

All methods use linear classifiers trained with Adam using a learning rate of $10^{-4}$.
Input features are standardized by z-score normalization computed on the training set, and the same statistics are applied to validation and test data.
Strategic manipulation costs are implemented using Euclidean and Mahalanobis distances, following standard practice in strategic classification.

Individual similarity is measured by standardized Euclidean distance.
For each dataset, the fairness threshold $\delta$ is calibrated according to empirical pairwise-distance quantiles.
For example, on the German dataset, $\delta\in\{4,5.2,5.9,6.8\}$ corresponds to the 5\%, 10\%, 15\%, and 20\% quantiles, respectively.
Peer-driven manipulation is instantiated using the three visibility regimes described in Section~\ref{sec:visibility}.
During training, the stochastic perturbation operator $\mathcal{T}(\cdot)$ replaces 20\% of visible peers with alternative positively decided peers to simulate visibility uncertainty.

We evaluate all methods under four manipulation regimes: classical manipulation, peer imitation, mixed manipulation, and noisy manipulation.
All results are averaged over 10 random seeds, with standard deviations reported.
Further details on cost functions, visibility instantiation, manipulation construction, and evaluation protocols are provided in Appendix~\ref{app:implementation}.

\begin{figure*}[t]
    \centering
    \begin{subfigure}[b]{0.239\textwidth}
        \includegraphics[width=\linewidth]{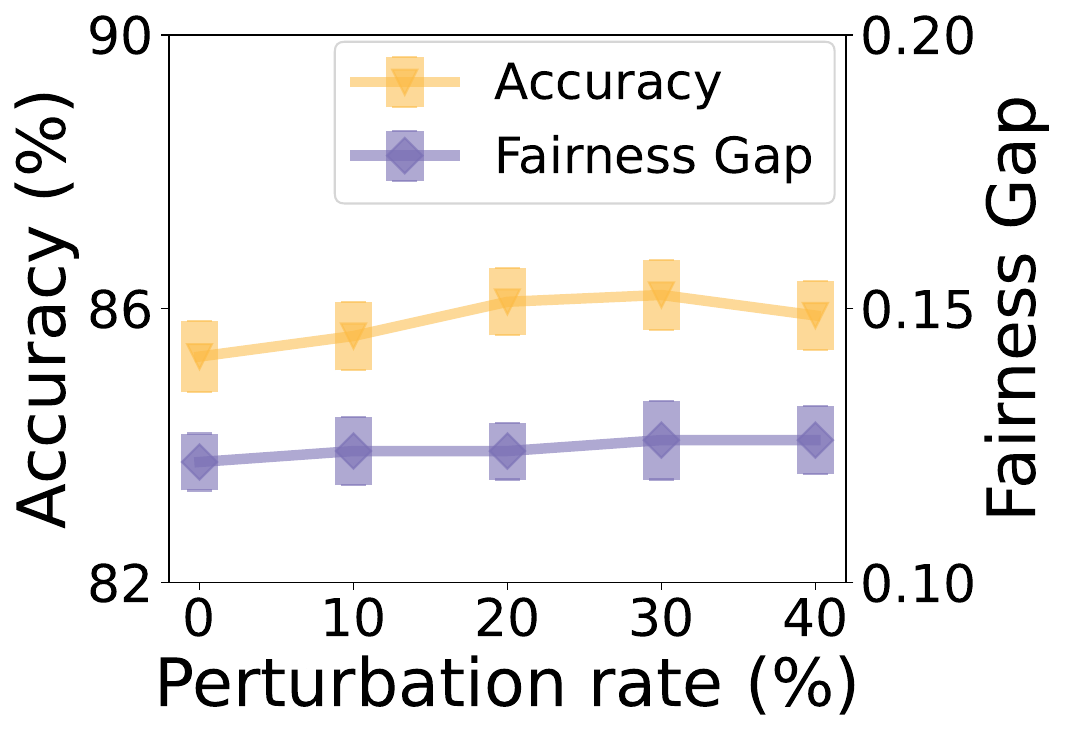}
        \caption{Test with Local visibility.}
        \label{fig8a}
    \end{subfigure}
    \begin{subfigure}[b]{0.239\textwidth}
        \includegraphics[width=\linewidth]{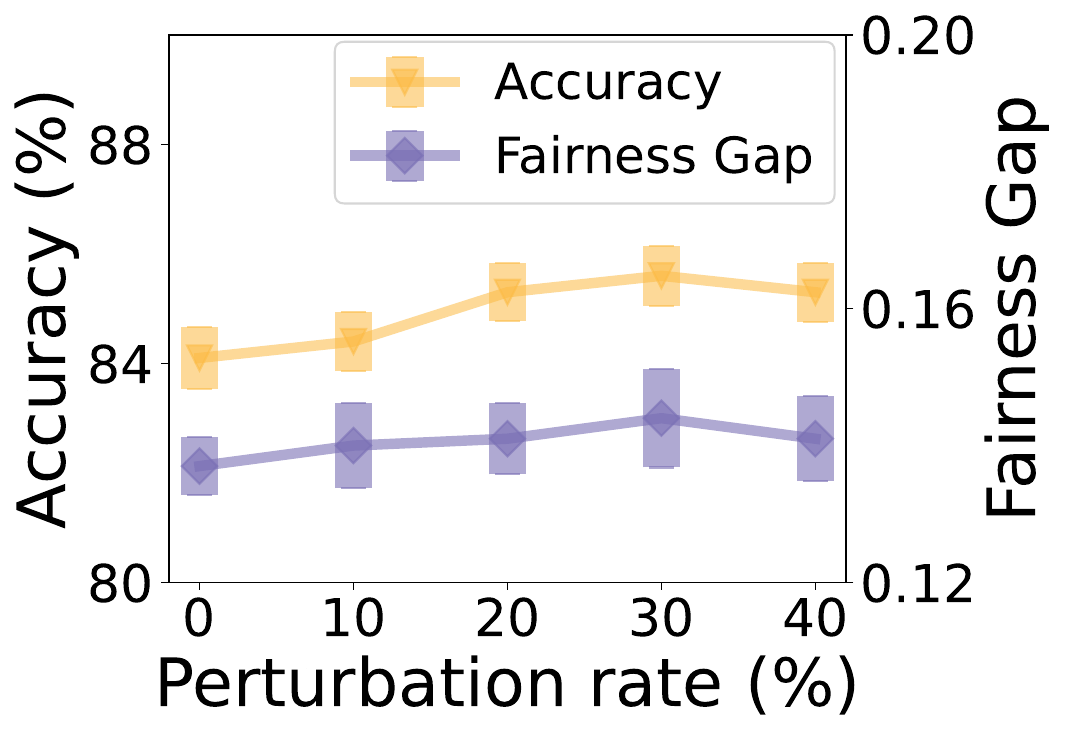}
        \caption{Test with Prob visibility.}
        \label{fig8b}
    \end{subfigure}
    \begin{subfigure}[b]{0.239\textwidth}
        \includegraphics[width=\linewidth]{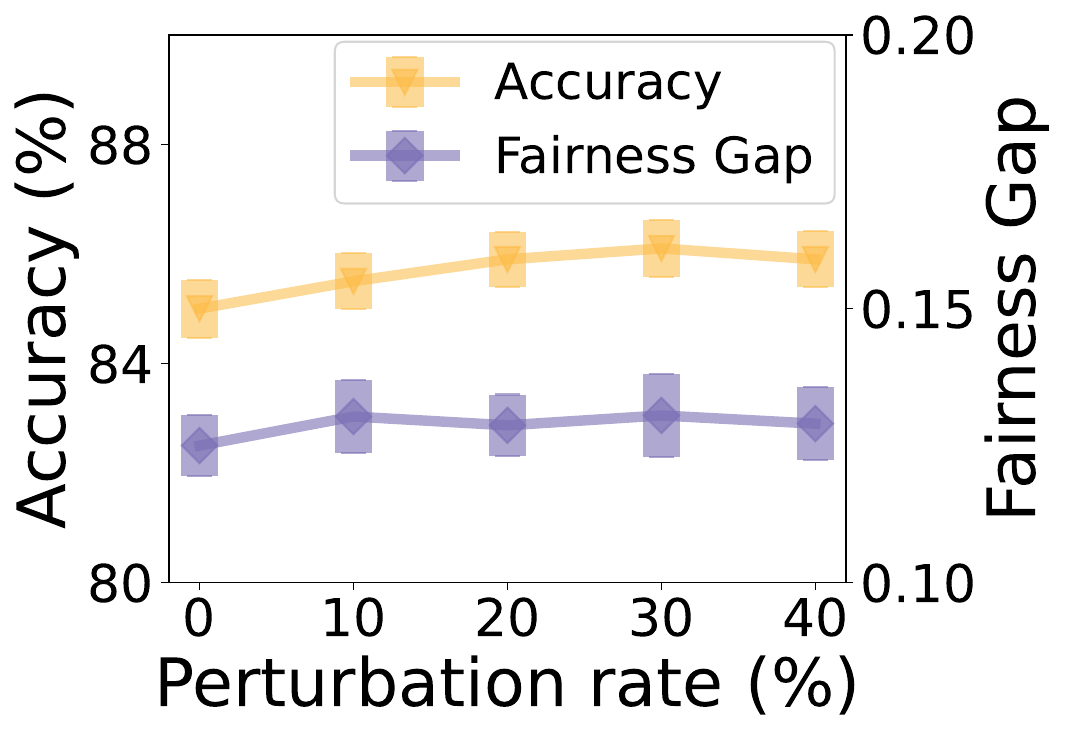}
        \caption{Test with Local visibility.}
        \label{fig8c}
    \end{subfigure}
    \begin{subfigure}[b]{0.239\textwidth}
        \includegraphics[width=\linewidth]{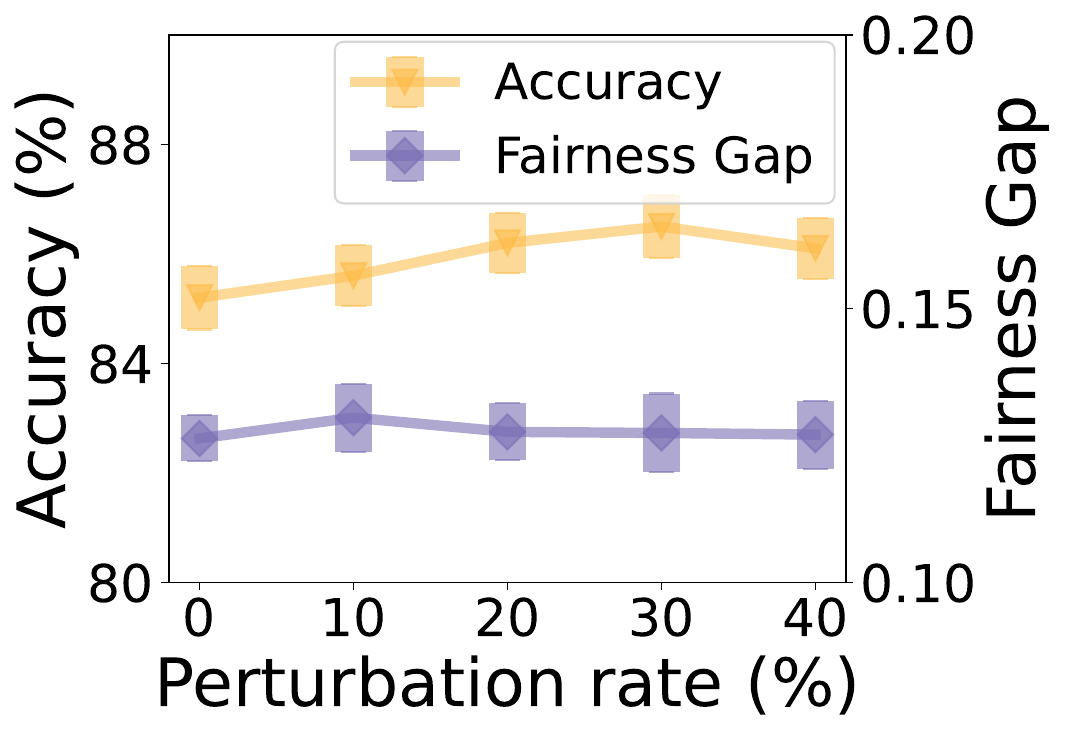}
        \caption{Test with kNN visibility.}
        \label{fig8d}
    \end{subfigure}
    \vskip -0.06in
    \caption{Effect of stochastic perturbation on post-manipulation accuracy and individual fairness gap under visibility mismatch. The left two panels correspond to models trained with kNN visibility, and the right two panels correspond to models trained with Prob visibility.}
    \label{fig8}
\end{figure*}

\subsection{Results and Analysis}
\noindent \textbf{Peer-driven manipulation vs. independent manipulation.}
Figure~\ref{fig2} compares independent manipulation with our three peer-driven manipulations in terms of manipulation effectiveness, measured by the cheatment rate.
Across both Mahalanobis and Euclidean manipulation costs, peer-driven manipulation yields consistently higher cheatment rates than independent manipulation.
Specifically, the effectiveness gap widens as the individual-fairness threshold $\delta$ increases: when $\delta$ is small, imitation is restricted to a tight similarity neighborhood and the advantage over independent manipulation is limited, whereas larger $\delta$ allows agents to imitate a broader set of similar accepted peers, resulting in higher cheatment rates.
This result suggests that under individual fairness constraints, agents can exploit peer imitation to manipulate more effectively.

\noindent \textbf{Post-manipulation accuracy and individual fairness.}
Tables~\ref{tab:acc_main} and~\ref{tab:ifgap_main} jointly evaluate post-manipulation accuracy and individual fairness under different manipulation regimes.
Under Standard SC, classifiers trained under independent manipulation assumptions degrade when agents adopt peer imitation behavior, typically resulting in a $2$--$4\%$ drop in accuracy together with increased decision inconsistency among $\delta$-similar individuals; a milder degradation (around $1$--$2\%$) is observed under mixed manipulation.
Notably, peer imitation alone may partially reduce the individual fairness gap under Standard SC, as agents move toward existing positively decided regions, which locally aligns decisions within similarity neighborhoods.
In contrast, IFSC consistently achieves higher post-manipulation accuracy while reducing the individual fairness gap across datasets.
The improvement persists under mixed manipulation, indicating that incorporating peer imitation responses during training reduces the mismatch between assumed and realized strategic behavior.
Overall, IFSC remains accurate and individually fair even when the deployed population exhibits heterogeneous strategic responses.

\noindent \textbf{Robustness under visibility mismatch.}
Figure~\ref{fig5} evaluates post-manipulation accuracy and individual fairness gap when the visibility regime used during training differs from that at test time. Classifiers trained under a fixed visibility assumption exhibit noticeable performance degradation under mismatched visibility regimes, manifested as reduced accuracy and increased fairness gap. In contrast, IFSC maintains more stable performance across visibility settings, achieving consistently higher accuracy and lower fairness gap. This indicates that explicitly modeling peer imitation during training improves robustness to variations in observable peer sets and reduces sensitivity to visibility assumptions.

\noindent \textbf{Effect of stochastic perturbation.}
Figure~\ref{fig8} illustrates the effect of stochastic perturbation during training under visibility mismatch. As the perturbation rate increases, post-manipulation accuracy becomes more stable across different test visibility regimes, while the individual fairness gap remains largely unchanged with only minor fluctuations. These results suggest that stochastic perturbation primarily enhances robustness to visibility variations without compromising individual fairness.

\section{Conclusion}

We investigate strategic classification under individual fairness and show that classical agent-independent response models fail to capture the strategic interdependence induced by fairness constraints.
This reveals a fundamental mismatch between existing strategic classification formulations and fairness-aware decision settings.
To address this issue, we model peer-driven manipulation arising from individual fairness and develop a learning framework that explicitly accounts for the resulting dependence among strategic responses.
Experiments on synthetic and real-world datasets demonstrate that the proposed approach improves robustness to strategic behavior while maintaining strong predictive performance and individual fairness after manipulation. Future work includes extending the framework to more complex strategic interaction models and studying long-term population dynamics.

\section*{Acknowledgement}
This work was jointly supported by the National Natural Science Foundation of China~(Nos.~62276004, 62325604, 62525213), Tsinghua
University-Siemens Joint Research Center~(JCIIOT), the Shanghai Sailing Program~(No.~24YF2711600), the Natural Science Foundation of Heilongjiang Province~(Grant No.~LH2023C069), the NUDT Youth Independent Innovation Science Fund~(No. ZK25-20), and the NUDT Innovation Foundation for Postgraduate~(No.XJQY2025011).

\bibliographystyle{plainnat}
\bibliography{sample-base}

\appendix

\section{Proof of Proposition~\ref{prop:gf_not_imply_if}}
\label{appB}
We provide a constructive counterexample showing that satisfying the group fairness constraint in~\eqref{eq:group_fair_sc} does not guarantee the individual fairness requirement in~\eqref{eq:if_sc_post}.

\textbf{Define a strategic setting where post-manipulation features coincide with pre-manipulation features.}
Consider a strategic classification instance in which manipulation is (effectively) absent: for every agent, the best response is the identity map, so that $\mathbf{x}'=\mathbf{x}$ almost surely.
This can be realized, for example, by taking a manipulation cost that makes any nontrivial change infeasible or strictly suboptimal, so the induced post-manipulation distribution $\mathcal{D}'$ equals the original distribution $\mathcal{D}$ in terms of features.

\textbf{Construct a distribution with equal group-level positive rates.}
Let the feature space be $\mathcal{X}=\mathbb{R}$ and define the similarity metric as $d(u,v)=|u-v|$.
Fix any radius $\delta>0$ and choose two feature points
\begin{equation}
\mathbf{u}=0,
\qquad
\mathbf{v}=\tfrac{\delta}{2},
\end{equation}
so that $d(\mathbf{u},\mathbf{v})=\delta/2\le \delta$.

Consider two groups $G=\{0,1\}$ and define a distribution $\mathcal{D}$ over $(\mathbf{x},g)$ as follows:
for each group $g\in G$,
\begin{equation}
\Pr(\mathbf{x}=\mathbf{u}\mid g)=\tfrac12,
\qquad
\Pr(\mathbf{x}=\mathbf{v}\mid g)=\tfrac12.
\end{equation}
In other words, within \emph{each} group, half of the mass is at $\mathbf{u}$ and half at $\mathbf{v}$, and the two groups have identical feature distributions.

\textbf{Define a classifier that satisfies group fairness but violates individual fairness.}
Define a (deterministic) classifier $f$ by
\begin{equation}
f(\mathbf{u})=1,
\qquad
f(\mathbf{v})=0.
\end{equation}
Then for each group $g\in G$,
\begin{equation}
\Pr\big(f(\mathbf{x}')=1\mid g\big)
=
\Pr\big(f(\mathbf{x})=1\mid g\big)
=
\Pr(\mathbf{x}=\mathbf{u}\mid g)
=
\tfrac12,
\end{equation}
where we used $\mathbf{x}'=\mathbf{x}$ from Step~1.
Hence the group-conditioned positive prediction rates are identical across groups.

Now consider the group fairness constraint in~\eqref{eq:group_fair_sc}.
For common disparity functionals $\Phi$ used in group fairness (e.g., maximum difference across groups, variance across groups, or any functional that equals $0$ when all group rates are identical), we obtain
\begin{equation}
\Phi\Big(\{\Pr(f(\mathbf{x}')=1\mid g)\}_{g\in G}\Big)=0.
\end{equation}
Therefore, the constraint in~\eqref{eq:group_fair_sc} is satisfied for any tolerance $\varepsilon\ge 0$.

However, individual fairness in~\eqref{eq:if_sc_post} is violated: since
\begin{equation}
d(\mathbf{u},\mathbf{v}) \le \delta
\quad\text{but}\quad
f(\mathbf{u})\neq f(\mathbf{v}),
\end{equation}
there exists a pair of $\delta$-similar individuals receiving different decisions.
Thus~\eqref{eq:if_sc_post} fails.

We have constructed $\mathcal{D}$, $d$, $\delta$, and $f$ such that the group fairness constraint in~\eqref{eq:group_fair_sc} holds while the individual fairness requirement in~\eqref{eq:if_sc_post} is violated.

\section{Proof of Proposition~\ref{pro1}}
\label{app:proof_pro1}

We prove~\eqref{eq:peer_not_separable} by contradiction using a population-switching argument.

Fix any similarity radius $\delta>0$ and consider a feature space $\mathcal{X}=\mathbb{R}$ with metric $d(u,v)=|u-v|$.
Fix a decision rule $f:\mathcal{X}\to\{0,1\}$ that is not constant.
Then there exist two points $a,c\in\mathcal{X}$ such that
\begin{equation}
f(a)=1
\qquad\text{and}\qquad
f(c)=0.
\end{equation}
Fix an agent index $i$ with feature $\mathbf{x}_i=c$.
Let $b_{\mathrm{IF}}(\mathbf{x}_i; f, \{\mathbf{x}_j\}_{j=1}^n)$ denote a peer-driven manipulation rule, meaning that whenever peer imitation is feasible for agent $i$ under a given population realization, the output $\mathbf{x}'_i$ satisfies the peer imitation condition in~\eqref{eq:peer_imitation_def}.

\noindent\textbf{Construct two populations with identical $(\mathbf{x}_i,f)$.}
We construct two population realizations $P$ and $P'$ of size $n$ such that agent $i$ has the same feature in both populations, i.e., $\mathbf{x}_i=\mathbf{x}'_i=c$, and the deployed rule $f$ is identical, but the locations of positively decided peers differ.

\noindent\textbf{Population $P$:} include at least one peer located at $a$.
That is, there exists an index $j^\star\neq i$ such that $\mathbf{x}_{j^\star}=a$.

\noindent\textbf{Population $P'$:} ensure that no positively decided peer lies within the $\delta$-neighborhood of $a$.
Formally, for every $j$ such that $f(\mathbf{x}'_j)=1$, we require $d(\mathbf{x}'_j,a)>\delta$.

Both populations share the same agent feature $\mathbf{x}_i=c$ and the same decision rule $f$.

\noindent\textbf{Peer imitation feasibility differs across populations.}
In population $P$, the point $a$ is present and is positively decided ($f(a)=1$).
Thus, peer imitation is feasible for agent $i$: for instance, choosing $\mathbf{x}'_i=a$ satisfies~\eqref{eq:peer_imitation_def} with the witness peer $j^\star$.

In population $P'$, by construction there is no positively decided peer within distance $\delta$ of $a$.
Therefore, any post-manipulation feature that lies in the $\delta$-neighborhood of $a$ (including $a$ itself) cannot satisfy~\eqref{eq:peer_imitation_def} in population $P'$, since there is no positively decided peer that can serve as the required reference point.
Hence, the set of post-manipulation features that satisfy the peer imitation condition~\eqref{eq:peer_imitation_def} differs between $P$ and $P'$.

\noindent\textbf{Derive a contradiction for any independent manipulation function.}
Assume for contradiction that there exists an \emph{independent} manipulation function $b(\cdot;f)$ such that
\begin{equation}
b_{\mathrm{IF}}\!\left(\mathbf{x}_i; f, \{\mathbf{x}_j\}_{j=1}^n\right)
\equiv
b(\mathbf{x}_i; f)
\quad\text{for all population realizations.}
\end{equation}
Since $\mathbf{x}_i=c$ and $f$ are identical under populations $P$ and $P'$, the output of the independent function must be the same under both populations:
\begin{equation}
b(\mathbf{x}_i;f)=b(\mathbf{x}'_i;f).
\end{equation}
However, by Step~2, the peer imitation feasibility structure differs across $P$ and $P'$.
In particular, there exist post-manipulation points (e.g., $a$ and points in its $\delta$-neighborhood) that satisfy~\eqref{eq:peer_imitation_def} under $P$ but cannot satisfy it under $P'$.
Thus, no single population-independent output $b(\mathbf{x}_i;f)$ can coincide with a peer-driven manipulation across both populations while respecting~\eqref{eq:peer_imitation_def} whenever it is feasible.

This contradicts the assumed equivalence between $b_{\mathrm{IF}}(\mathbf{x}_i; f, \{\mathbf{x}_j\}_{j=1}^n)$ and $b(\mathbf{x}_i;f)$.
Therefore, in general,
\begin{equation}
    b_{\mathrm{IF}}\!\left(\mathbf{x}_i; f, \{\mathbf{x}_j\}_{j=1}^n\right)\ \not\equiv\ b(\mathbf{x}_i; f),
\end{equation}
which proves~\eqref{eq:peer_not_separable}.

\section{Experimental Details}
\label{app:implementation}

\subsection{Implementation Details}

\noindent\textbf{Classifier and optimization.}
All classifiers are implemented as linear logistic models.
Model parameters are optimized using Adam with a learning rate $10^{-4}$ and default momentum parameters.
Training is performed for 100 epochs on real-world datasets and 200 epochs on the synthetic dataset.
The model with the lowest validation loss is selected for evaluation.
All input features are standardized using z-score normalization computed from the training split only.

\noindent\textbf{Manipulation cost models.}
We consider two manipulation cost functions.
The Euclidean cost is defined as
\begin{equation}
c_{\mathrm{Euc}}(\mathbf{x}_i,\mathbf{x}'_i)
=
\|\mathbf{x}'_i-\mathbf{x}_i\|_2 .
\end{equation}
The Mahalanobis cost is defined as
\begin{equation}
c_{\mathrm{Mah}}(\mathbf{x}_i,\mathbf{x}'_i)
=
\sqrt{
(\mathbf{x}'_i-\mathbf{x}_i)^\top
\mathbf{M}
(\mathbf{x}'_i-\mathbf{x}_i)
},
\end{equation}
where $\mathbf{M}\succeq 0$ is set to the inverse empirical covariance matrix of the training features.
This cost captures feature correlations and heterogeneous manipulation difficulty.

\noindent\textbf{Individual fairness metric.}
Individual similarity is measured by standardized Euclidean distance on normalized features.
The fairness threshold $\delta$ is calibrated using empirical pairwise-distance quantiles for each dataset.
For example, on the German dataset, $\delta\in\{4,5.2,5.9,6.8\}$ corresponds to the 5\%, 10\%, 15\%, and 20\% quantiles of pairwise distances, respectively.
The same calibration strategy is applied to other datasets.

\noindent\textbf{Visibility model instantiation.}
For metric-based local visibility, the visibility radius $R_{\mathrm{vis}}$ is selected from empirical distance quantiles so that each agent observes a small local set of positively decided peers.
For salient-feature kNN visibility, similarity is computed on a salient feature subset, and each agent observes the $k$ nearest positively decided peers in this subspace.
For probabilistic visibility, each positively decided peer is sampled according to the distance-decay kernel in Section~\ref{sec:visibility}, with the bandwidth $\sigma$ set proportional to the median pairwise distance among positively decided individuals.
The same visibility parameters are used across methods for fair comparison.

\noindent\textbf{Peer-driven manipulation simulation.}
Given a deployed classifier $f$, we first identify the positively decided peer set $S^{+}(f)$.
For each agent $i$, the visible peer set $S_i^{\mathrm{vis}}(f)$ is then constructed according to one of the visibility regimes.
The agent selects a visible positively decided peer as its reference and searches for a feasible post-manipulation feature within the corresponding $\delta$-neighborhood that maximizes its utility under the specified manipulation cost.
This produces the peer-driven response defined in Eq.~\eqref{eq:if_manipulation}.

\noindent\textbf{Visibility-robust perturbation.}
During training, we perturb the visible peer set to simulate uncertainty in peer observability.
Given $S_i^{\mathrm{vis}}(f)$, the perturbed set
$\widetilde{S}_i^{\mathrm{vis}}(f)\sim\mathcal{T}(S_i^{\mathrm{vis}}(f))$
is constructed by replacing 20\% of visible peers with alternative positively decided peers sampled from $S^{+}(f)$.
The resulting manipulated features $\widetilde{\mathbf{x}}'_i$ are used for robust classifier training.

\noindent\textbf{Manipulation regimes.}
We evaluate classifiers under four manipulation regimes.
Classical manipulation follows the standard independent best-response model in Eq.~\eqref{eq:best_response}.
Peer imitation follows the peer-driven response model in Eq.~\eqref{eq:if_manipulation}.
Mixed manipulation assigns each agent to classical manipulation or peer imitation with equal probability.
Noisy manipulation is used as a distribution-mismatched test setting: agents start from the classical best response, but the resulting post-manipulation feature is perturbed by zero-mean Gaussian noise and clipped to the valid feature range.
This regime is not explicitly used by either Standard SC or IFSC during training, and therefore provides an additional robustness evaluation.

\noindent\textbf{Peer-driven mixture regime.}
To simulate heterogeneous peer imitation, we combine visibility models and cost functions.
Let
\begin{equation}
\mathcal{V}=\{v_{\mathrm{local}},v_{\mathrm{kNN}},v_{\mathrm{prob}}\}
\quad\text{and}\quad
\mathcal{C}=\{c_{\mathrm{Euc}},c_{\mathrm{Mah}}\}.
\end{equation}
We define the component set as $\mathcal{G}=\mathcal{V}\times\mathcal{C}$.
For each agent $i$, a component $g_i=(v_i,c_i)$ is sampled uniformly from $\mathcal{G}$ and fixed throughout evaluation.
The agent then performs peer imitation using the selected visibility model and manipulation cost.

\noindent\textbf{Evaluation protocol.}
All experiments are repeated over 10 random seeds.
Reported results are means with standard deviations.
Train, validation, and test splits are kept fixed across methods under the same random seed to ensure fair comparison.

\end{document}